\pdfoutput=1

\documentclass[11pt]{article}

\usepackage[preprint]{acl}
\usepackage{acl}
\usepackage{times}
\usepackage{latexsym}
\usepackage[T1]{fontenc}

\usepackage[utf8]{inputenc}

\usepackage{microtype}

\usepackage{inconsolata}

\usepackage{graphicx}
\usepackage{subcaption}
\usepackage{hyperref} 
\usepackage{amsmath}
\usepackage{booktabs}
\title{Refining Packing and Shuffling Strategies for Enhanced Performance in Generative Language Models}

\author{
 \textbf{Yanbing Chen}\textsuperscript{2}\thanks{*These authors contributed equally to this work.},
 \textbf{Ruilin Wang}\textsuperscript{3}\footnotemark[1],
  \textbf{Zihao Yang}\textsuperscript{3},
 \textbf{Lavender Yao Jiang}\textsuperscript{1,3},
 \textbf{Eric Karl Oermann}\textsuperscript{1,3}\thanks{†Corresponding author.} 
\\
 \textsuperscript{1}Department of Neurosurgery, NYU Langone Health,\\
 \textsuperscript{2}School of Global Public Health, New York University,\\
 \textsuperscript{3}Center for Data Science, New York University,
\\
 \small{
   \textbf{Correspondence:} 
   \href{mailto:yc6785@nyu.edu}{yc6785@nyu.edu},
   \href{mailto:rw2608@nyu.edu}{rw2608@nyu.edu},
   \href{mailto:gavin.yang@nyu.edu}{gavin.yang@nyu.edu},
   \href{mailto:lyj2002@nyu.edu}{lyj2002@nyu.edu},
   \href{mailto: eric.oermann@nyulangone.org}{eric.oermann@nyulangone.org}
 }
}

\begin{document}
\maketitle
\begin{abstract}

Packing and shuffling tokens is a common practice in training auto-regressive language models (LMs) to prevent overfitting and improve efficiency. Typically documents are concatenated to chunks of maximum sequence length (MSL) and then shuffled. However setting the atom size, the length for each data chunk accompanied by random shuffling, to MSL may lead to contextual incoherence due to tokens from different documents being packed into the same chunk. An alternative approach is to utilize padding, another common data packing strategy, to avoid contextual incoherence by only including one document in each shuffled chunk. To optimize both packing strategies (concatenation vs padding), we investigated the optimal atom size for shuffling and compared their performance and efficiency. We found that matching atom size to MSL optimizes performance for both packing methods (concatenation and padding), and padding yields lower final perplexity (higher performance) than concatenation at the cost of more training steps and lower compute efficiency. This trade-off informs the choice of packing methods in training language models~\footnote{The codebase available on github: \url{https://anonymous.4open.science/r/data-shuffling-3A4D/README.md}}.
\end{abstract}

\section{Introduction}\label{sec:intro}

Dataset shuffling removes underlying chronological or thematic order in the original dataset, which reduces the risk of overfitting and improves model generalizability \citep{nicolae2016towards,shen2020magnet,zhong2023rinas}. For example, training a classifier for cats versus dogs with a dataset containing 5,000 images of each can lead to bias if the dataset is not shuffled. If the first 5,000 gradient updates are solely from cat images, the model develops a "cat bias," making inference on dogs problematic. This issue can be avoided by interleaving cat and dog images, and this process of shuffling datasets prior to training machine learning models has become a standard approach.

Although shuffling facilitates unbiased learning by providing independent samples, the optimal packing approach for data shuffling in generative language model remains unclear  \citep{press2019partially,abdou2022word}.  For GPT models \citep{radford2019language}, the commonly used PyTorch Dataloader class concatenates and packs documents into chunks of a fixed size (usually MSL) before shuffling (hereafter referred to as 'concat'). Another approach, the padding method, shuffles documents after padding them to a fixed size. Both methods achieve the goal of generating fixed-length sequences, but it is still an open question which method is more effective for GPT models. 

In terms of packing methods for language models, the appropriate shuffling unit size remains uncertain. For some modeling tasks, like the visual classification mentioned above, we can shuffle the training data in units of one or a few images. However, since training datasets for language models consist of documents or sequences with varying lengths, selecting the appropriate shuffling unit size is challenging. For conciseness in later discussions, we define the unit of data length used in the shuffling process as "atom size".



We hypothesize that shuffling data in an atom size of MSL is best since the contextual information within each shuffling chunk is maximized. Specifically, transformers approximate the next token distribution given the previous context. This context would be disrupted when the atom size is smaller than the MSL, as unrelated contextual fragments are concatenated together. In addition, the context of consecutive sequences would be dependent when the atom size is larger than the MSL, thus introducing correlation and bias. Therefore, using the MSL as the shuffling unit preserves both integrity and randomness of the training data. 


Our experiments confirmed that packing and shuffling data in atom sizes of MSL optimizes performance for both concat and padding methods. We also showed that padding results in better model performance than concat, albeit at the cost of efficiency due to more training steps. 

\section{Method}\label{sec:method}

\subsection{Model Pretraining Setting}\label{sec:model_pretraining_setting}
We pretrained GPT-2 124M models~\citep{radford2019language} on WikiText with each packing method—concat or padding—across various atom sizes and MSLs. Table~\ref{tab:choices} shows different configurations tested. We used Alibi\citep{press2021train}  as a positional encoding that introduces no additional learnable parameters, such that all models have the same total parameter size regardless of their MSLs. One observation in padding is that models have different step sizes, which is discussed in Appendix~\ref{sec:step_size_diff_in_padding}. Models were trained for 2 epochs on 1 NVIDIA A100 GPU. Appendices~\ref{sec:dataset_and_filtering}, ~\ref{sec:concat_and_padding}, ~\ref{sec:total_parameter_size}, and ~\ref{sec:total_step_size} provide details on dataset and filtering, specific implementations for data packing methods, an explanation for parameter sizes with Alibi, and an justification for the step size, respectively.

\subsection{Evaluation and Comparison Metric}\label{sec:evaluation_and_comparison_metric}
We used final perplexity and perplexity ranking to determine the optimal atom size for both packing methods (concat or padding) across 3 MSLs, resulting in 28 experiments overall. Detailed calculations for final perplexity and perplexity ranking are explained in Appendix~\ref{sec:calculation_of_perplexity}. Under each MSL, we compared concat and padding models by final perplexity, learning efficiency (perplexity at given steps) and step size efficiency (steps per epoch). Table~\ref{tab:choices} lists MSL and atom size choices, with justifications provided in Appendix~\ref{sec:choice_of_msl_and_atom_size}.


\begin{table}[htbp]
\centering
\resizebox{\linewidth}{!}{%
\begin{tabular}{lcc}
\toprule
\textbf{MSL} & \textbf{Atom Size Choice for Both Concat and Padding.} \\
\midrule
32 & 8, 16, 32, 64, 128 \\
64 & 16, 32, 64, 128, 256 \\
128 & 32, 64, 128, 256 \\
\bottomrule
\end{tabular}%
}
\caption{MSL and atom size choices}
\label{tab:choices}
\end{table}

\section{Results}\label{sec:result} 
\subsection{Concat Experiments}\label{sec:concat_experiment}
\begin{figure*}[t]
    \centering
    \begin{subfigure}[b]{0.45\textwidth} 
        \includegraphics[width=\textwidth]{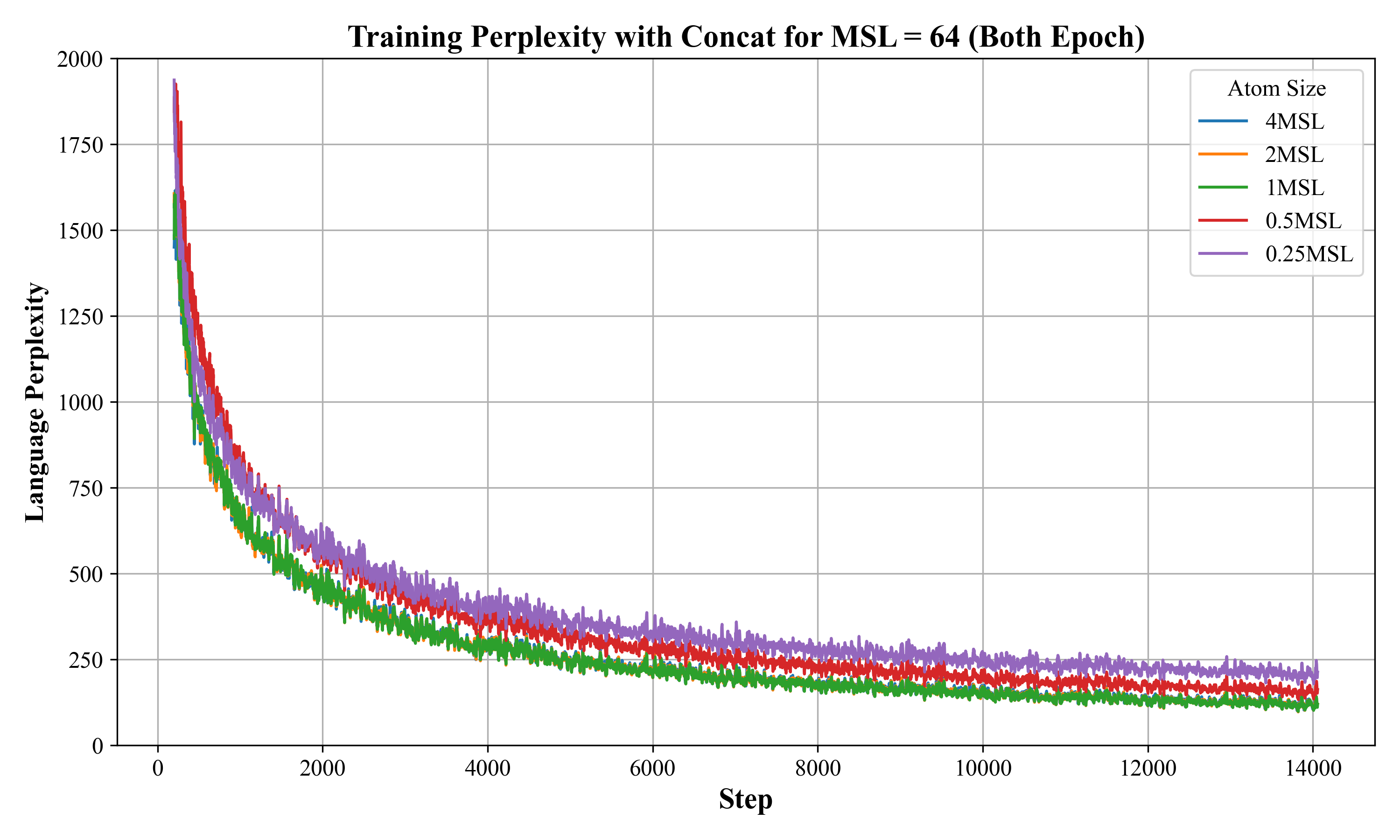}
        \caption{\textbf{Full Training Perplexity. }The models with atom sizes of 0.5MSL (red) and 0.25MSL (purple) have higher perplexity than the others. 1MSL (green) stabilizes at a low perplexity after an initial drop.}
  
    \end{subfigure}
    \hfill
    \begin{subfigure}[b]{0.45\textwidth}
        \includegraphics[width=\textwidth]{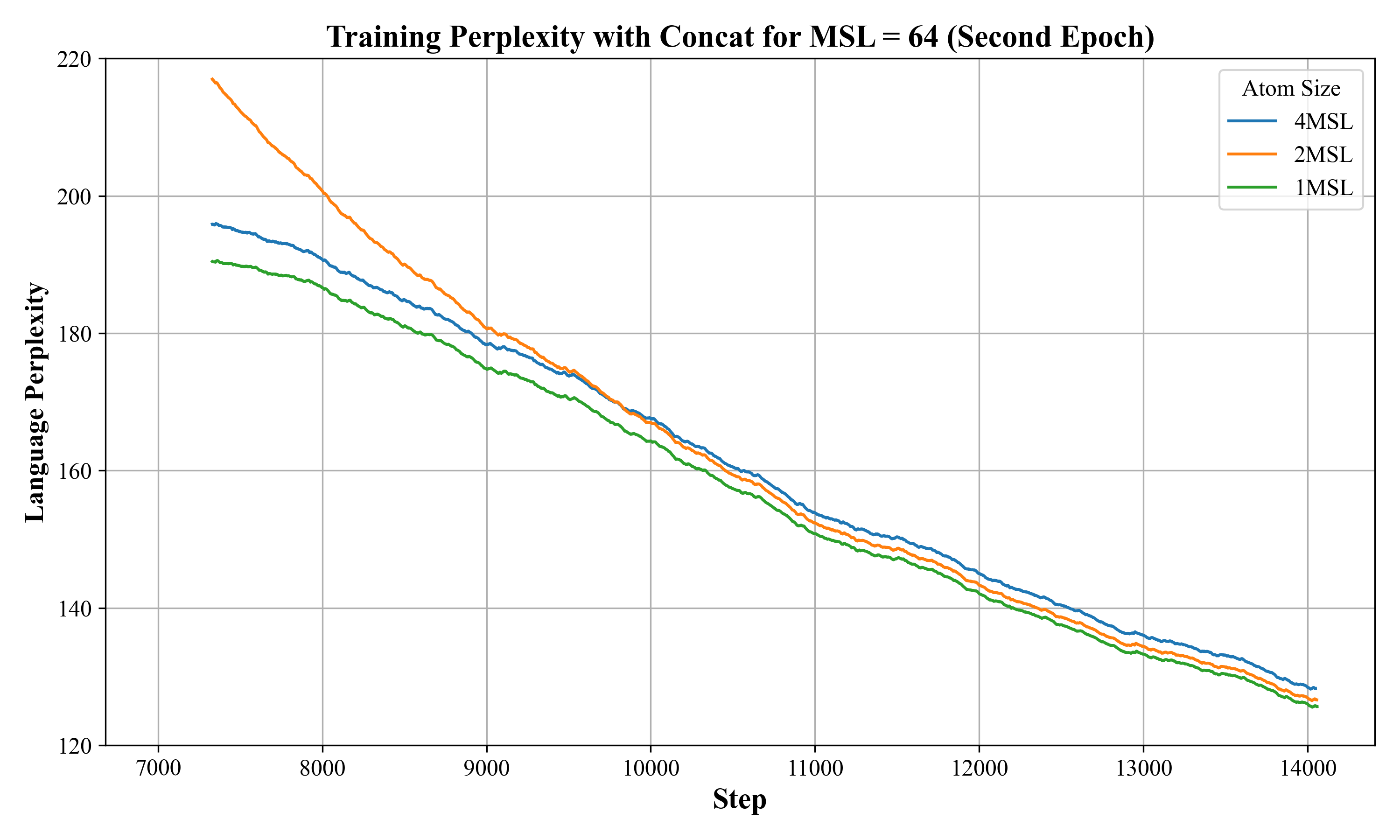}
        \caption{\textbf{Second Epoch Perplexity.} Initially, the model with atom size of 2MSL (orange) has higher perplexity than the other two. 1MSL (green) consistently maintains the lowest perplexity in the second epoch.}
    \end{subfigure}
    \caption{Comparisons across concat models with different atom sizes when MSL is 64. Smaller or larger atom sizes than 1MSL increase perplexity. The model with 1MSL as the atom size has the lowest final perplexity at the end of 2 epochs,  indicating the best performance.}
    \label{fig:concat_results}
\end{figure*}

We found that atom sizes smaller or larger than MSL increased perplexity, indicating that MSL is indeed the optimal atom size for concat. Figure~\ref{fig:concat_results}(a) shows the training perplexity of concat models with different atom sizes $s \in \{$0.25MSL, 0.5MSL, $1$MSL, $2$MSL, $4$MSL$\}$ when MSL is 64. Among all atom sizes, 0.25MSL (purple) and 0.5MSL (red) obviously lead to higher perplexity (worse performance). Although the differences in perplexity among 4MSL (blue), 2MSL (orange) and 1MSL (green) are minimal, 1MSL consistently had lower perplexity (better performance) than 2MSL and 4MSL. Figure~\ref{fig:concat_results}(b) shows the training perplexity of the second epoch as an example.  Table~\ref{tab:pad_concat_ppl} shows final perplexity and perplexity ranking respectively. The model using 1MSL as the atom size has the lowest final perplexity (118.08) and highest average ranking (1.05), indicating optimal performance. Experiments with MSL = 32, 128 yielded similar results, as detailed in Appendix~\ref{sec:32_128}.

\begin{figure*}[t]
    \centering
    \begin{subfigure}[b]{0.45\textwidth}
        \includegraphics[width=\textwidth]{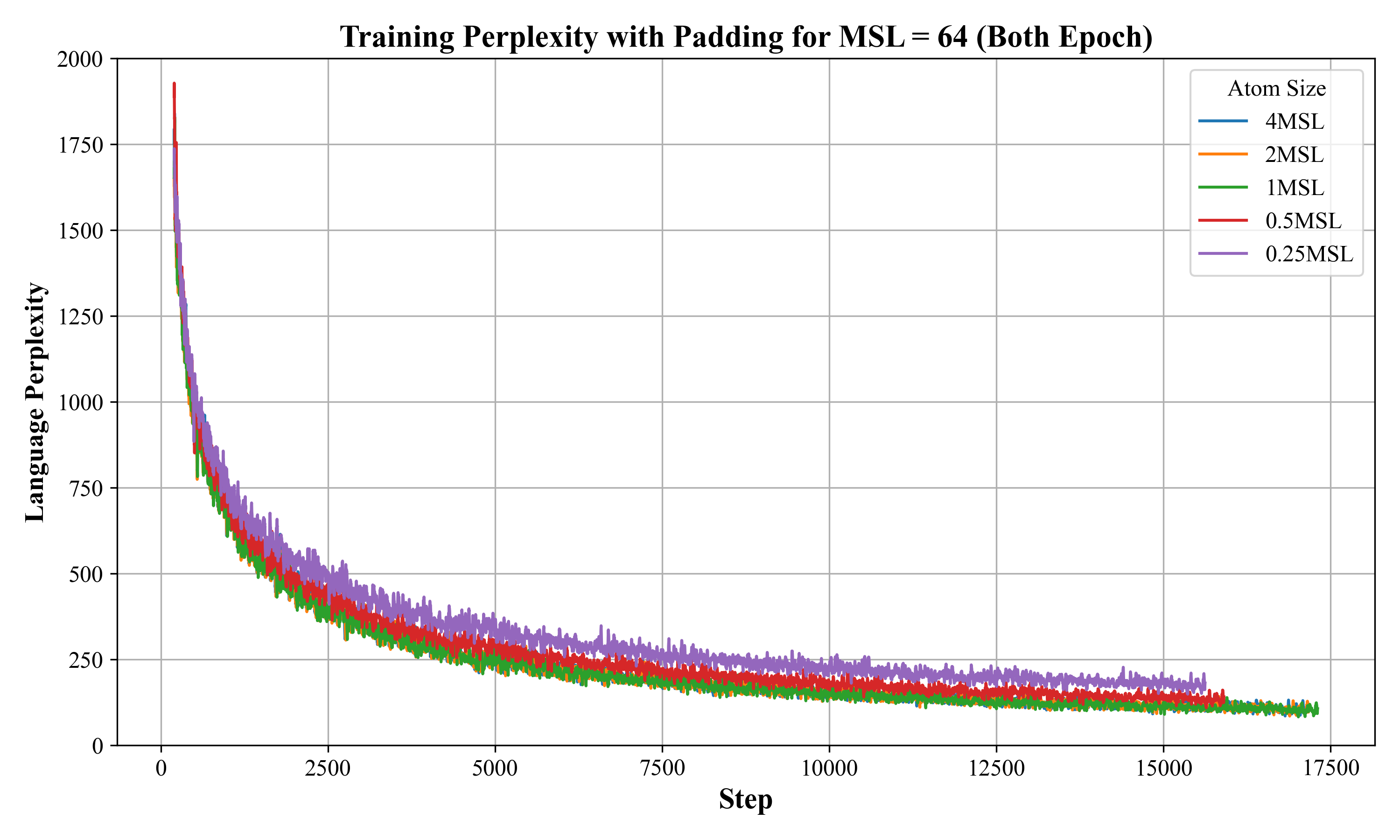}
        \caption{\textbf{Full Training Perplexity. }The models with atom sizes of 0.5MSL (red) and 0.25MSL (purple) have higher perplexity than the others. 1MSL (green) stabilizes at a low perplexity after an initial drop.}
    \end{subfigure}
    \hfill
    \begin{subfigure}[b]{0.45\textwidth}
        \includegraphics[width=\textwidth]{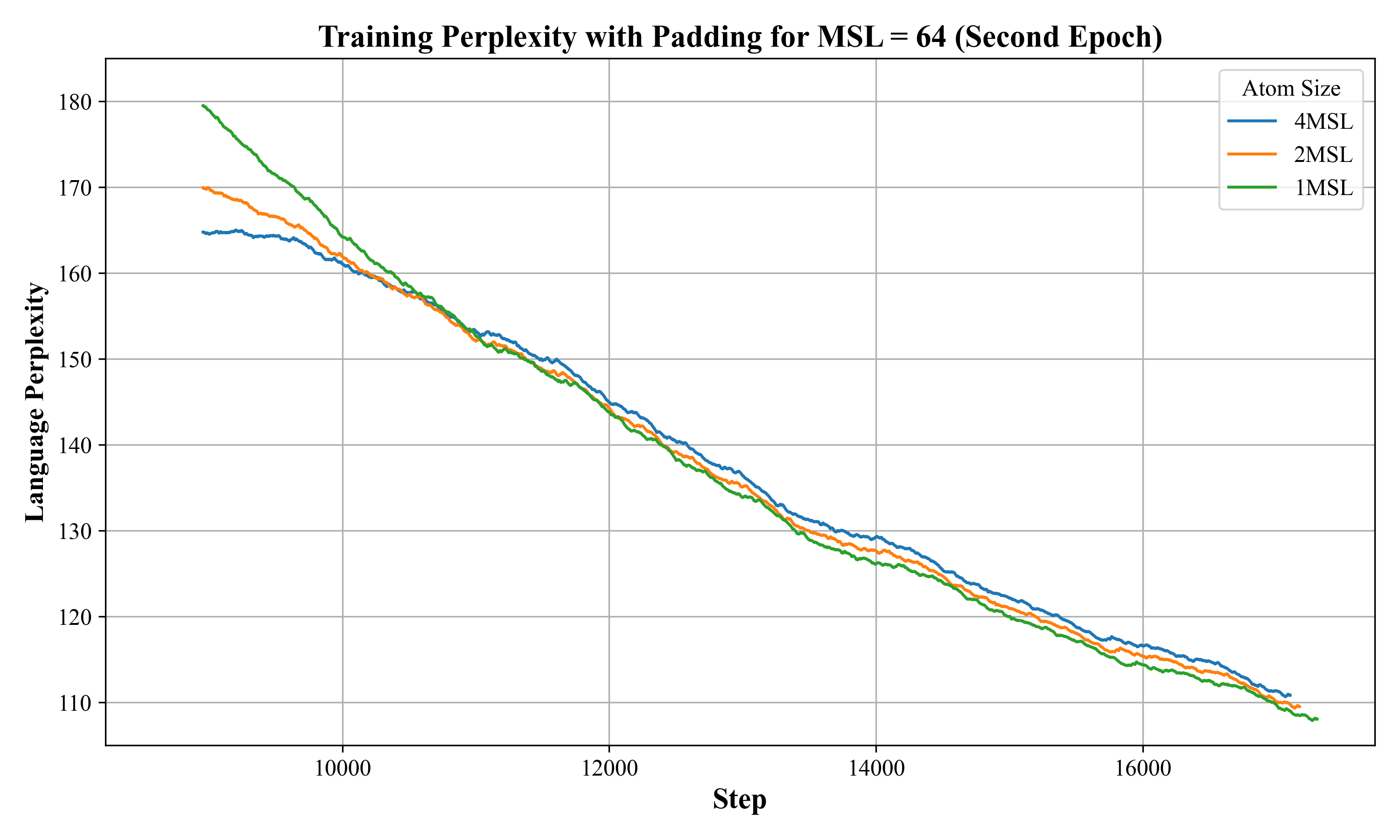}
        \caption{\textbf{Second Epoch Perplexity.} Initially, the model with atom size of 1MSL (green) shows higher perplexity than the other two. 1MSL continuously decreases and achieves the lowest perplexity by the end of the second epoch.}
    \end{subfigure}
    \caption{Comparisons across padding models with different atom sizes when MSL is 64. Smaller or larger atom sizes than MSL increase perplexity. The model with 1MSL as the atom size has the lowest final perplexity at the end of 2 epochs,  indicating the best performance.}
    \label{fig:padding_results}
\end{figure*}

\begin{table}[htbp]
\centering
\resizebox{\linewidth}{!}{%
\begin{tabular}{cccccc}
\toprule
\textbf{Atom Size} & \multicolumn{2}{c}{\textbf{Final Perplexity}} & \multicolumn{2}{c}{\textbf{Perplexity Ranking}} \\
& \textbf{Concat} & \textbf{Padding} & \textbf{Concat} & \textbf{Padding} \\
\midrule
0.25MSL& 207.04 & 175.33 & 5.00 & 5.00 \\
0.5MSL& 157.39 & 134.43 & 4.00 & 4.00 \\
1MSL& \textbf{118.08} & \textbf{102.82} & \textbf{1.05} & \textbf{1.18} \\
2MSL& 119.66& 104.46& 1.96& 2.03\\
4MSL& 121.18& 105.85& 2.99& 2.79\\

\bottomrule
\end{tabular}%
}
\vspace{-10pt}
\caption{Comparison of final perplexity values and average perplexity rankings across different atom sizes for concat and padding models when MSL is 64.}
\label{tab:pad_concat_ppl}
\end{table}

\subsection{Padding Experiments}\label{sec:padding_experiment}
For padding, we found that atom sizes smaller or larger than MSL increased perplexity, confirming MSL as the optimal atom size. Figure~\ref{fig:padding_results}(a) shows the training perplexity of padding models with atom sizes $s \in \{$0.25MSL, 0.5MSL, $1$MSL, $2$MSL, $4$MSL$\}$ when MSL is 64. It takes different training steps for different padding models to finish 1 epoch, as mentioned in Section~\ref{sec:step_size_diff_in_padding}.  Similar to the concat experiments, 0.25MSL (purple) and 0.5MSL (red) lead to higher perplexity, while differences between 4MSL (blue), 2MSL (orange) and 1MSL (green) are subtle. We found that 1MSL consistently had lower perplexity compared to 2MSL and 4MSL. Figure~\ref{fig:padding_results}(b) shows the second epoch's training perplexity, where 1MSL started with the highest perplexity but improved to perform better (lower perplexity) than 2MSL and 4MSL by the end of training. Table ~\ref{tab:pad_concat_ppl} presents the final perplexity and perplexity ranking. The model with atom size of 1MSL has the lowest final perplexity (102.82) and highest average ranking (1.18), indicating optimal performance. Experiments with MSL is 32 or 128 yielded similar results (See Appendix~\ref{sec:32_128}. for details). 

\subsection{Comparison between Padding and Concat}\label{sec:comparison_pad_concat}
Although the padding method resulted in lower final perplexities (better performance) than concat, it has lower learning efficiency (higher perplexity at given steps) and step size efficiency (more steps per epoch). Table~\ref{tab:pad_concat_compare} compares the total step size and final perplexity for concat and padding models when the atom size matches the MSL, showing that padding models have larger step sizes and lower final perplexity than concat models across MSLs.

\begin{figure*}[t]
    \centering
    \begin{subfigure}[b]{0.32\textwidth}
        \includegraphics[width=\textwidth]{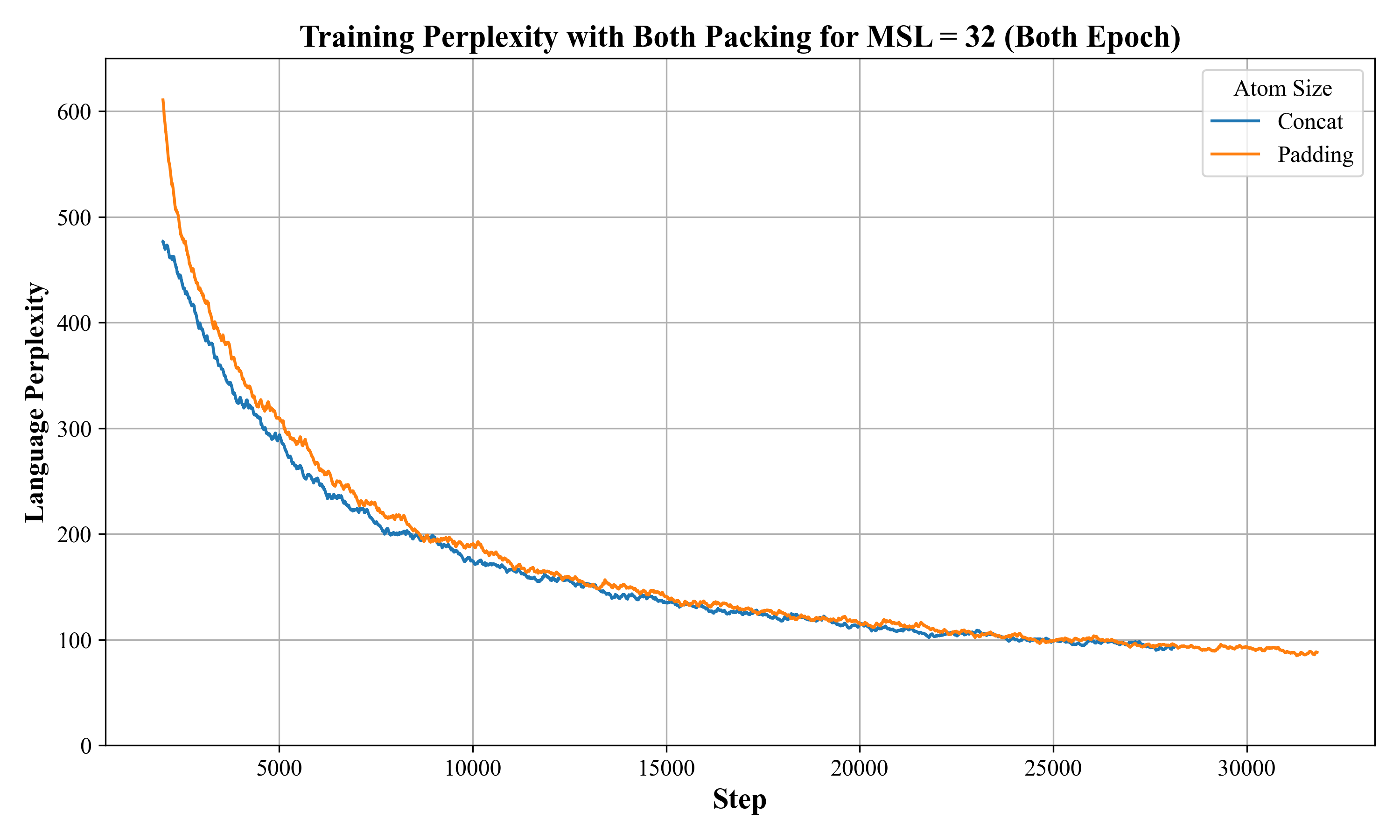}
        \caption{\textbf{MSL = 32.}}
    \end{subfigure}
    \begin{subfigure}[b]{0.32\textwidth}
        \includegraphics[width=\textwidth]{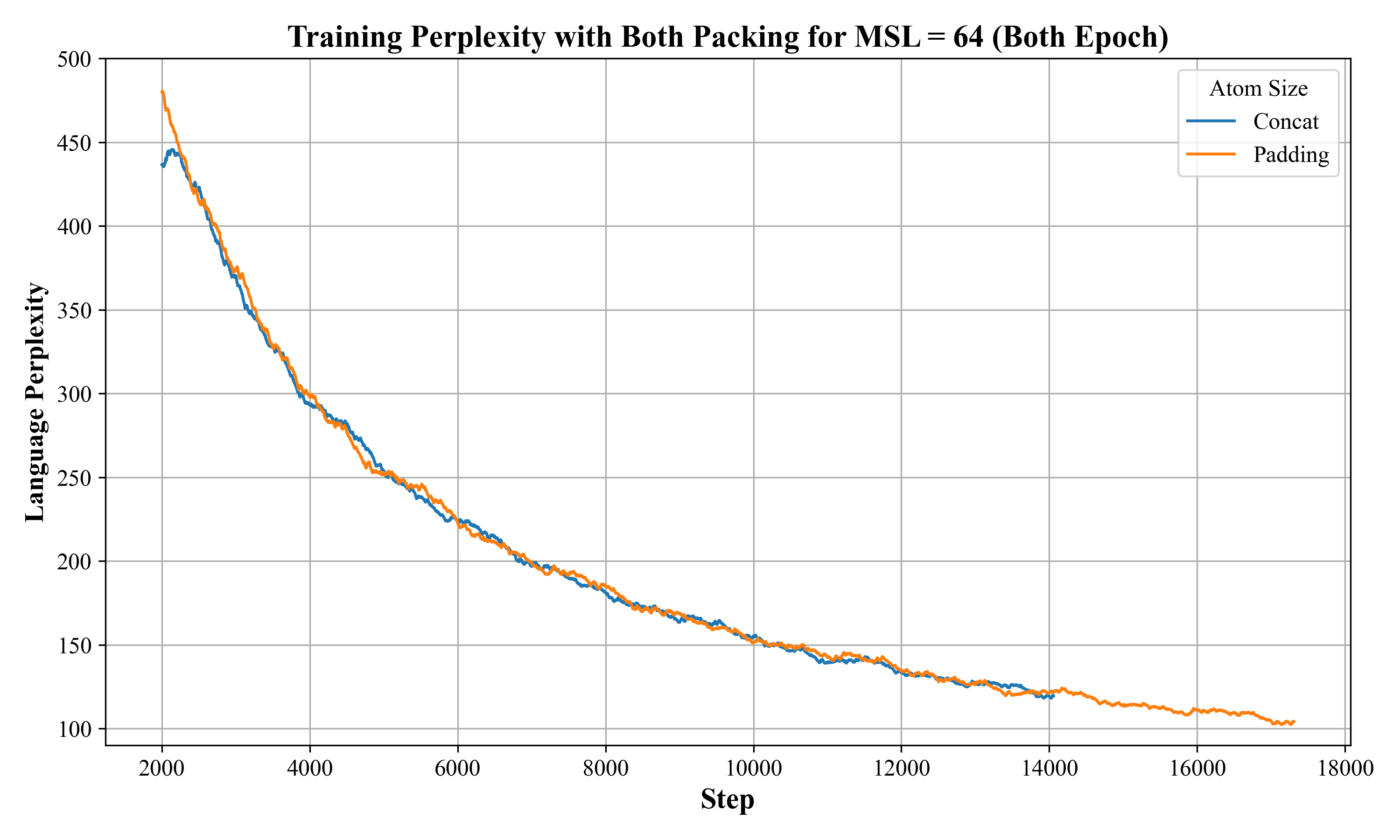}
        \caption{\textbf{MSL = 64.}}
    \end{subfigure}
    \begin{subfigure}[b]{0.32\textwidth}
        \includegraphics[width=\textwidth]{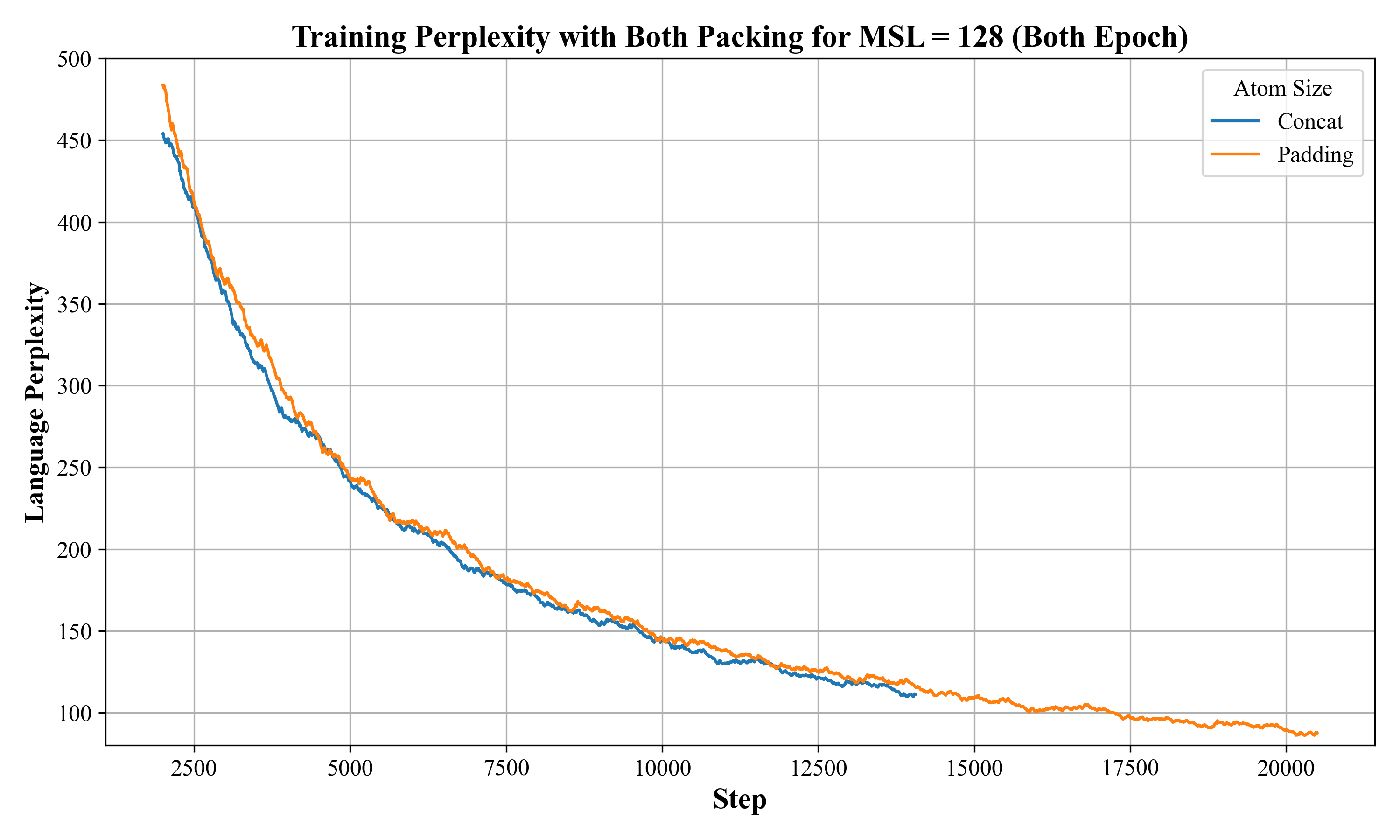}
        \caption{\textbf{MSL = 128.}}
    \end{subfigure}
    \caption{Step-wise comparison of perplexity between padding and concat models under different MSLs (the first 2,000 steps discarded due to high perplexity). Padding (orange) has lower final perplexities (better performance) while concat (blue) has smaller training step sizes over 2 epochs.}
    \label{fig:concat_vs_pad_ppl}
\end{figure*}

Additionally, Figure~\ref{fig:concat_vs_pad_ppl} shows the step-wise perplexity comparison for concat (blue) and padding (orange) models when the atom size matches the MSL (for clearer visualization, the first 2,000 steps are discarded due to high perplexity in all plots). Again, we see that padding has lower final perplexities while concat has smaller training step sizes.

\begin{table}[htbp]
\centering
\resizebox{\linewidth}{!}{%
\begin{tabular}{cccccc}
\toprule
\textbf{MSL} & \textbf{Batch Size} & \multicolumn{2}{c}{\textbf{Step Size}} & \multicolumn{2}{c}{\textbf{Final Step Perplexity}} \\
 & & \textbf{Concat} & \textbf{Padding} & \textbf{Concat} & \textbf{Padding} \\
\midrule
32 & 256 & \textbf{28120}& 31816 & 91.01 & \textbf{87.22}\\
64 & 256 & \textbf{14058}& 17308 & 110.45 & \textbf{99.79}\\
128 & 128 & \textbf{14056}& 20496 & 102.42 & \textbf{82.55}\\
\bottomrule
\end{tabular}%
}
\vspace{-10pt}
\caption{Comparison of total step size and final perplexity for concat and padding models under atom size = MSL, highlighting smaller step sizes and lower perplexities.}
\label{tab:pad_concat_compare}
\end{table}

\section{Discussion}\label{sec:discussion}
\subsection{Language Coherence and Bias}\label{sec:language_coherence__and_order_bias}
Matching MSL and atom size optimizes packing (padding and concat) by reducing language incoherence within a sequence and bias. Using an atom size smaller than MSL causes language incoherence, as it forces unrelated shuffling chunks to get merged into one sequence, damaging the contextual completeness of each sequence. Conversely,  atom size larger than MSL brings bias by splitting shuffling chunks across multiple consecutive sequences, creating unintended correlations between these sequences. 

\subsection{Implication of the Trade-off Between Performance and Efficiency}\label{sec:tradeoff_implication}
ML practitioners’ choice of packing methods may be informed by the trade-off between performance and efficiency. With limited amount of data, padding triumphs because it brings higher performance; with limited time, concat is preferable because it packs each epoch in fewer steps, leading to higher efficiency.

\section{Related Work}\label{sec:related_work}

\textbf{Shuffle in PyTorch.} While the DataLoader class in PyTorch shuffles data in concatenated chunks of a fixed atom size (usually MSL) to maximize training efficiency, our work explored multiple atom sizes (4MSL, 2MSL, 1MSL, 0.5MSL and 0.25MSL) as well as padding as an alternative packing method to optimize training performance.

\textbf{Data Shuffling Strategies for Context Preservation.} Zhao et al. \citep{zhao2024analysing} focused on intra-document causal attention mask as a packing strategy. In this strategy, documents are concatenated into chunks with fixed length, and the likelihood of each token is only conditioned on the previous tokens from the same document within the chunk. This is similar to padding because attention score is only calculated for intra-document tokens, but each token may not have full attention to other tokens in the same document since one document may be packed into different chunks. This method improves efficiency by saving padding tokens, but may suffer from contextual incompleteness.

\section{Conclusion}\label{sec:conclusion}
Our experiments using different packing methods with different atom sizes and MSLs show that matching atom size with maximum sequence length (MSL) optimizes packing performance (concat and padding). This finding underscores the importance of aligning atom size with MSL during data shuffling to optimize language model training.

We also found that padding yields lower final perplexity (higher performance) than concat at the cost of more training steps and lower efficiency. This trade-off guides packing choices in training models: padding is preferable when data is scarce, while concat is preferable when time is limited. \\

\textbf{Limitations and Future Work.} 
\begin{itemize}
    \item Our initial exploration showed MSL as the optimal atom size for packing and shuffling in GPT-2 124M models trained on WikiText. Future work using datasets with longer document lengths and other model architectures will further extend our findings.

    \item Our preliminary findings show that padding optimizes performance with limited amount of data. Specifically, we set MSL smaller than document lengths to avoid large amounts of padding tokens. However, this approach might not be practical in all settings, prompting future studies to explore padding's efficacy when MSL exceeds document lengths.
\end{itemize}



\newpage
\bibliography{custom}

\appendix

\section{Appendix}
\label{sec:appendix}
\subsection{Dataset and Filtering}\label{sec:dataset_and_filtering}
\begin{figure}[t]
    \centering
    \includegraphics[width=\columnwidth]{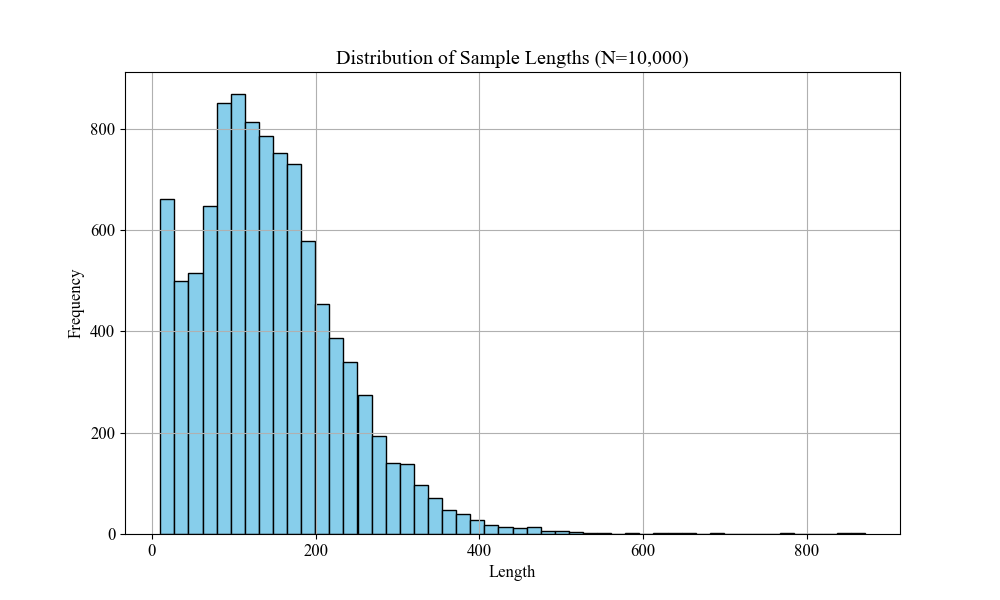} 
    \vspace{-10pt}
    \caption{The distribution of tokenized sequence lengths in WikiText-103-raw with 10,000 random samples. The dataset mostly consists of short paragraphs with length 0 to 200.}
    \label{fig:wiki}
\end{figure}

We conducted our studies on the generative language model training using the WikiText dataset \citep{merity2016pointer}, chosen for its generalizability.  Specifically, we used the WikiText-103-raw subset, which comprises approximately 1.81M rows and over 100M words derived from filtered Wikipedia content. Notably, the dataset mostly consists of short paragraphs: Figure~\ref{fig:wiki} shows the distribution of tokenized sequence lengths using 10,000 randomly sampled rows from the dataset. 

Before tokenization or shuffling the WikiText dataset, we removed blank rows and short title rows that contained limited context information. We filtered out rows with fewer than 50 words. This filtered 55.62\% rows (2.45\% words) in the training set and 53.86\% rows (2.33\% words) in the validation set. The training and validation corpus size after filtering are 98,937,698 and 208,893 words respectively.

Before feeding dataset to models, we preprocessed the sequences by tokenization and packing. We first used GPT2TokenizerFast~\citep{radford2019language}to tokenize all sequences in parallel, then used one of the two packing methods (padding and concat) with shuffling to ensure that all sequences could be batched in MSL. 

\subsection{Choices of MSL and Atom Size Explained}\label{sec:choice_of_msl_and_atom_size}
We set MSL = 32, 64, 128 to keep it smaller than document lengths and save wasteful padding tokens. Atom sizes were chosen to follow a geometric progression relative to MSL, set at 0.25MSL, 0.5MSL, 1MSL, 2MSL, 4MSL. However, when MSL = 128, we did not test on an atom size of 4 MSL = 512 because of wasteful padding tokens.
 
We also adjusted batch sizes based on MSL: 256 for MSLs of 32 and 64, and 128 for MSL = 128. These batch size selections were made to optimize GPU memory usage.

\begin{figure}[t]
    \centering
     \includegraphics[width=\columnwidth]{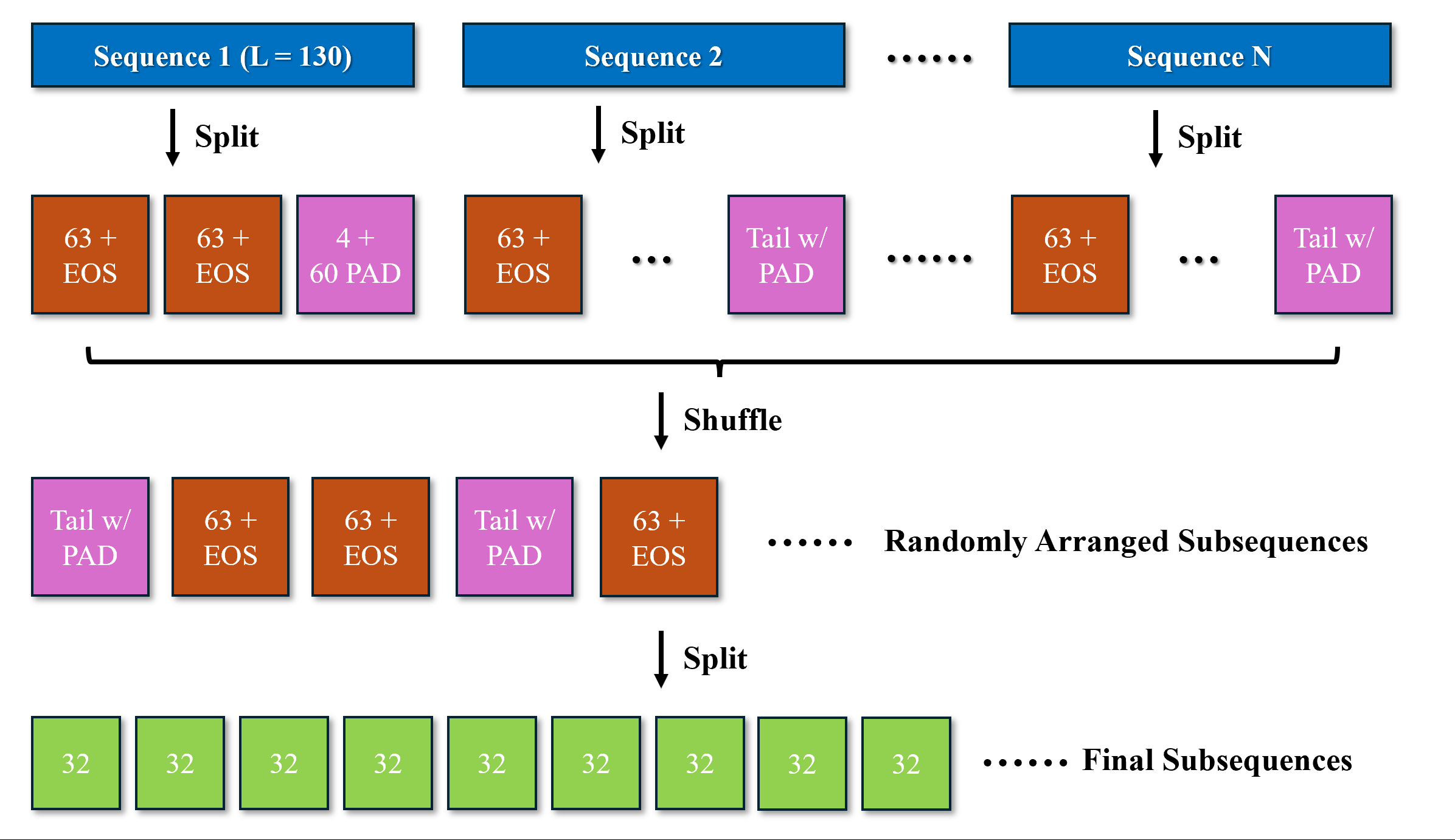} 
    \vspace{-10pt}
    \caption{Illustration of packing steps of padding, when MSL is 32 and atom size is 64. The "tail" subsequence contains fewer tokens than the specified atom size and is padded to meet the MSL requirement, ensuring consistency in sequence length.}
    \label{fig:padding}
\end{figure}
\subsection{Concat and Padding Details}\label{sec:concat_and_padding}
\textbf{Padding.}
This method focuses on padding to generate sequences with lengths equal to MSL. The steps are shown in Figure~\ref{fig:padding}. Each input document was segmented into smaller subsequences of length (atom size - 1) with an <EOS> token placed at the end. The role of the <EOS> token is to inform the model that the current sequence has ended. To maintain consistency in sequence length and ensure efficient batch processing, the tail end of any subsequence that does not meet the requirement of MSL would be padded. For example, in the case of MSL = 64, a sequence of length 130 ($L = 130$) would be segmented into 2 subsequences of length 64 (each with 63 word tokens and an <EOS> token at the end), then a tail subsequence composed of 4 word tokens and 60 padding tokens. We used <EOS> as the padding token for simplicity of the special token set. All resulted subsequences have a length of MSL regardless of the original sequence length.

Next, all subsequences were randomly shuffled with random seed set to 42. During this process, any underlying chronological or thematic order in the original dataset should be removed. 

After shuffling, the subsequences were either merged or split to align with the predefined MSL. When atom size is less than MSL, we merged subsequences; when atom size is larger than MSL, we split subsequences. Finally, when atom size equals MSL, we kept the shuffled subsequences unchanged. For example, in the case where MSL is 32 and atom size is 64, we split every subsequences to get 2 final subsequences of length 32 to feed to the model.

Notably, when atom size is larger than MSL, we do not pad every tail end to atom size, but to MSL instead as shown in figure~\ref{fig:padding}. This is because all subsequences will be split into size of MSL after shuffling. If we pad tail end to atom size instead of MSL, we will produce some training sequences that are completely composed of padding tokens. For example, when MSL is 32 and atom size is 128, if we pad a document of 35 tokens to an atom size of 128, we will yield a subsequence with 35 word tokens and 93 padding tokens, which will lead to completely meaningless training samples after split.

\begin{figure}[t]
    \centering
    \includegraphics[width=\columnwidth]{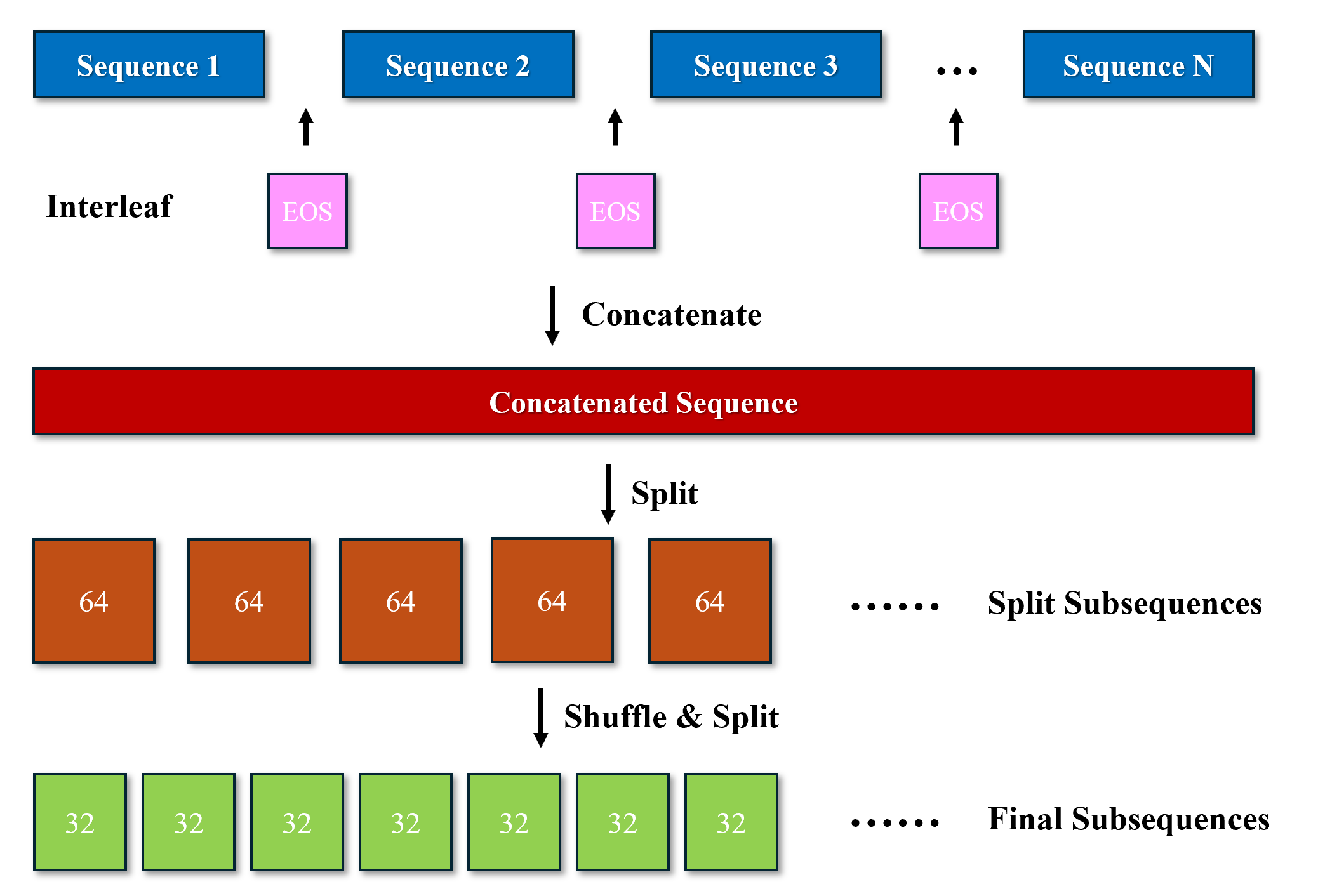} 
    \vspace{-10pt}
    \caption{Illustration of packing steps of concat, with MSL of 32 and atom size of 64.}
    \label{fig:method2}
\end{figure}

\textbf{Concat.}
While the padding method handles different sequence lengths with padding tokens, the concat method employs a concatenating and splitting process. The steps are shown in Figure~\ref{fig:method2}. In this approach, we firstly concatenate all sequences together to obtain an extremely long sequence interleaved with <EOS> tokens. Then we split the long sequence into subsequences according to atom size, shuffled them,  and adjusted them to fit the maximum context length by merging or splitting as needed. 

The two methods reflect different strategies to achieve the goal of generating fixed-length sequences for model training. The default method for GPT models is concat, but we hope to test whether padding outperforms concat in terms of efficiency and model performance. Our experiments in Section ~\ref{sec:result} would show that padding had better model performance than concat with the cost of lower efficiency. Since our experiments on shuffling atom size would focus on optimizing model performance, we decided to employ padding during those experiments.  

\subsection{Total Parameter Size}\label{sec:total_parameter_size}
When we experiment on models with different MSL, it is important to control the parameter size across all models. In the vanilla GPT-2 Small architecture, different MSLs lead to different parameter sizes. This is  because the positional encoding layer's parameter size has a positive linear relationship with maximum context length. To eliminate this difference, we decide to replace the positional encoding with Attention with Linear Biases (Alibi) \citep{press2021train}, which is a non-parametric positional encoding algorithm that biases attention scores in accordance with the distance between tokens. This algorithm was originally designed to improve the processing of long sequences in language models and to reduce the computational load associated with longer inputs. However, our smaller-scale model with shorter maximum context lengths did not benefit from these advantages. Instead, we focused on one particular characteristic of Alibi: it does not introduce additional trainable parameters, unlike the default positional encoding in the GPT-2 architecture. As a result, the total parameter size of all models trained in our experiments were fixed to 124M.

\subsection{Total Step Size}\label{sec:total_step_size}
Our objective is to schedule and use computational resources in ways that minimize training time and adhere to optimal training practices. We estimate the optimal number of tokens for training based on the estimation table from Chinchilla \citep{hoffmann2022training}. The table indicates that a model with 400M parameters requires 8B tokens, so our 124M-parameter model will need  $(124/400)\times8 = 2.48$B tokens. Then we divided this number by batch size and maximum context length to calculate the optimal step size for training. However, due to limitations in computational resources, it would take us one day to train one model on the optimal number of tokens. In this case, we decided to run two full epochs (114,400,095 word tokens per epoch) for all experiments instead, bringing the perplexity to nearly convergence in 2 to 3 hours on one GPU (NVIDIA Tesla V100-PCIE-32GB ). 

\subsection{Detailed Calculation of Final Perplexity and Perplexity Ranking }\label{sec:calculation_of_perplexity}
After evaluating with perplexity, we compared all models based on their average perplexity ranking and final perplexity value. Here, we need to be careful of how to calculate these two comparison metrics in detail.

We chose to calculate the comparison metrics by epochs rather than training steps due to variations in the number of word tokens learned at each step in padding models. For ranking, we divided the last epoch into 100 segments, each covering 0.01 of an epoch, calculated the average perplexity for all models within each segment, and ranked them. We then averaged the rankings from all 100 ranges as our final ranking. For the final perplexity value, we selected the last range (0.99 epoch - 1.00 epoch) and calculated its average perplexity.



\subsection{Exponential Moving Average (EMA) }\label{sec:smoothing}
We use Exponential Moving Average (EMA) to visualize smooth perplexity curves for our models. EMA computes a weighted average of past data points, with exponentially decreasing weights that effectively smooth out fluctuations in original perplexity values. Specifically, EMA is computed iteratively using the following formula:

\[ S_t = \alpha \cdot y_t + (1 - \alpha) \cdot S_{t-1} \]

where \( S_t \) represents the smoothed perplexity at step \( t \), \( y_t \) is the observed perplexity at step \( t \), and \( \alpha \) is the smoothing parameter. \( \alpha \) is designed to dynamically adjust based on changes in training steps \( \Delta t \):

\[ \alpha_t = \min(\sqrt{\alpha}, 0.999)^{\Delta t} \]

where \( \Delta t \) is always 1 since training step is a discrete variable. Therefore, the smoothing parameter remains constant during the process. 
 
\subsection{Concat and Padding Results under MSL = 32 and 128 }\label{sec:32_128}
\begin{figure*}[t]
    \centering
    \begin{subfigure}[b]{0.45\textwidth}
        \includegraphics[width=\textwidth]{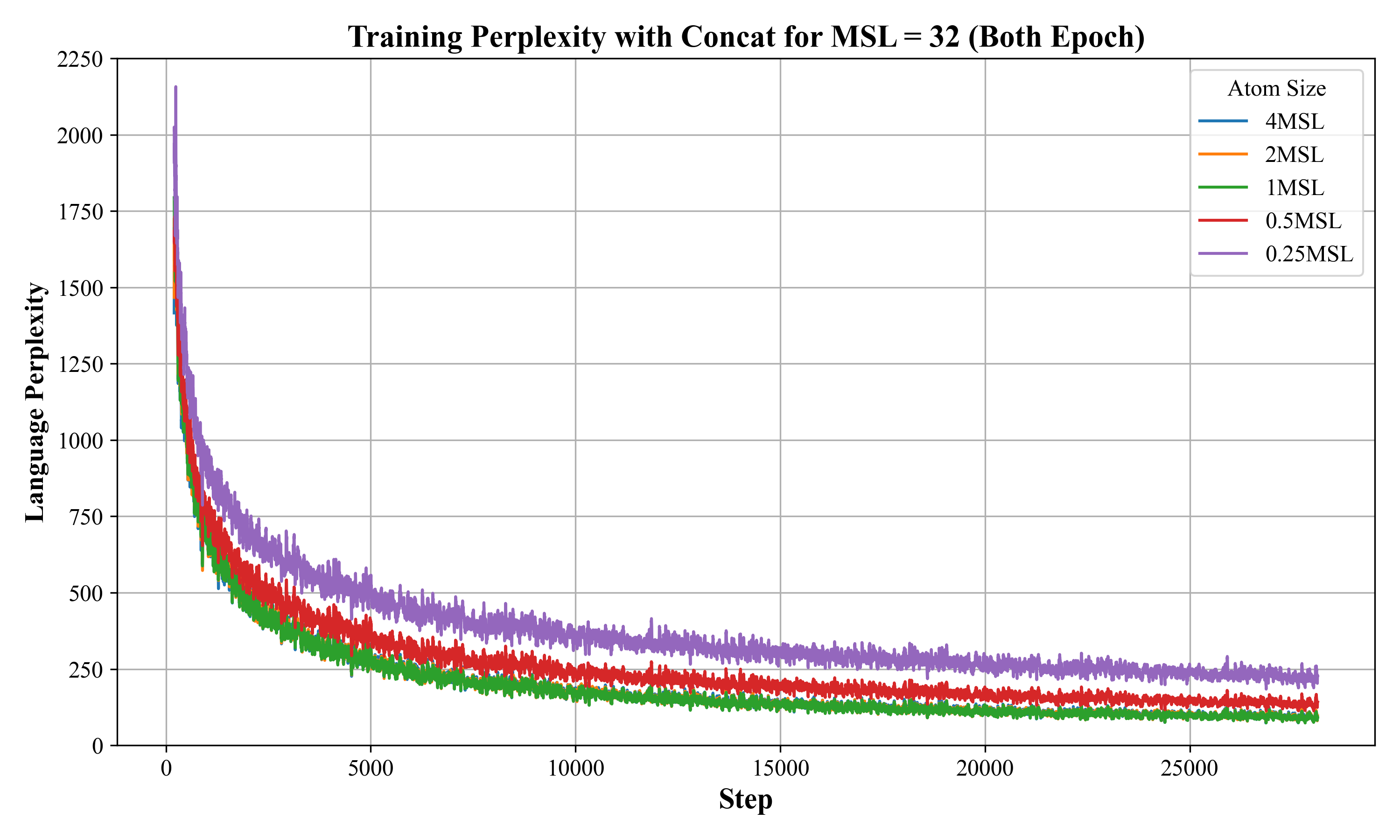}
        \caption{\textbf{Full Training Perplexity.} The models with atom sizes of 0.5MSL (red) and 0.25MSL (purple) have higher perplexity than the others. 1MSL (green) stabilizes at a low perplexity after an initial drop.}
    \end{subfigure}
    \hfill
    \begin{subfigure}[b]{0.45\textwidth}
        \includegraphics[width=\textwidth]{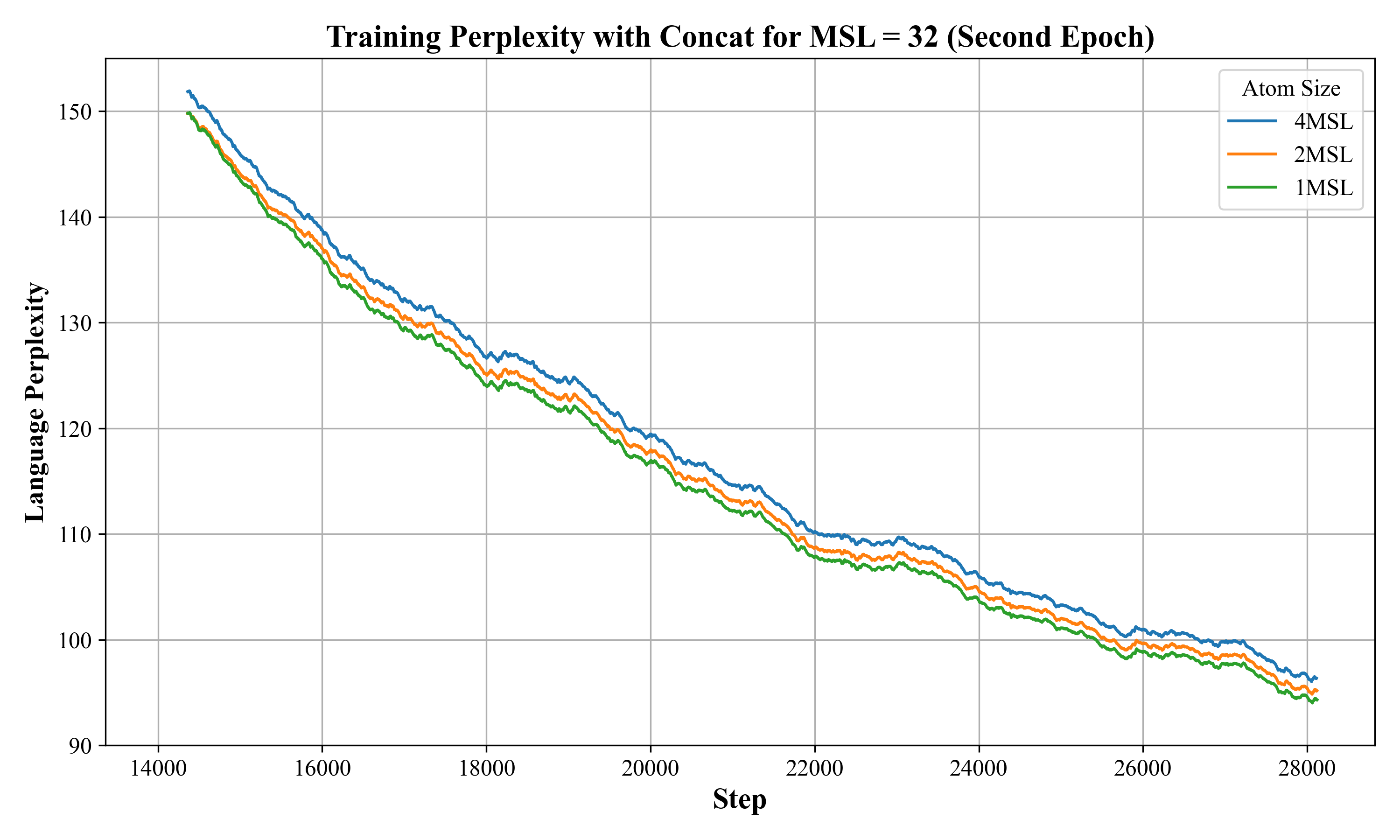}
        \caption{\textbf{Second Epoch Perplexity.} Models with atom size of 4MSL (blue) has higher perplexity than the other two. 1MSL (green) consistently maintains the lowest perplexity in the second epoch.}
    \end{subfigure}
    \vskip\baselineskip
    \begin{subfigure}[b]{0.45\textwidth}
        \includegraphics[width=\textwidth]{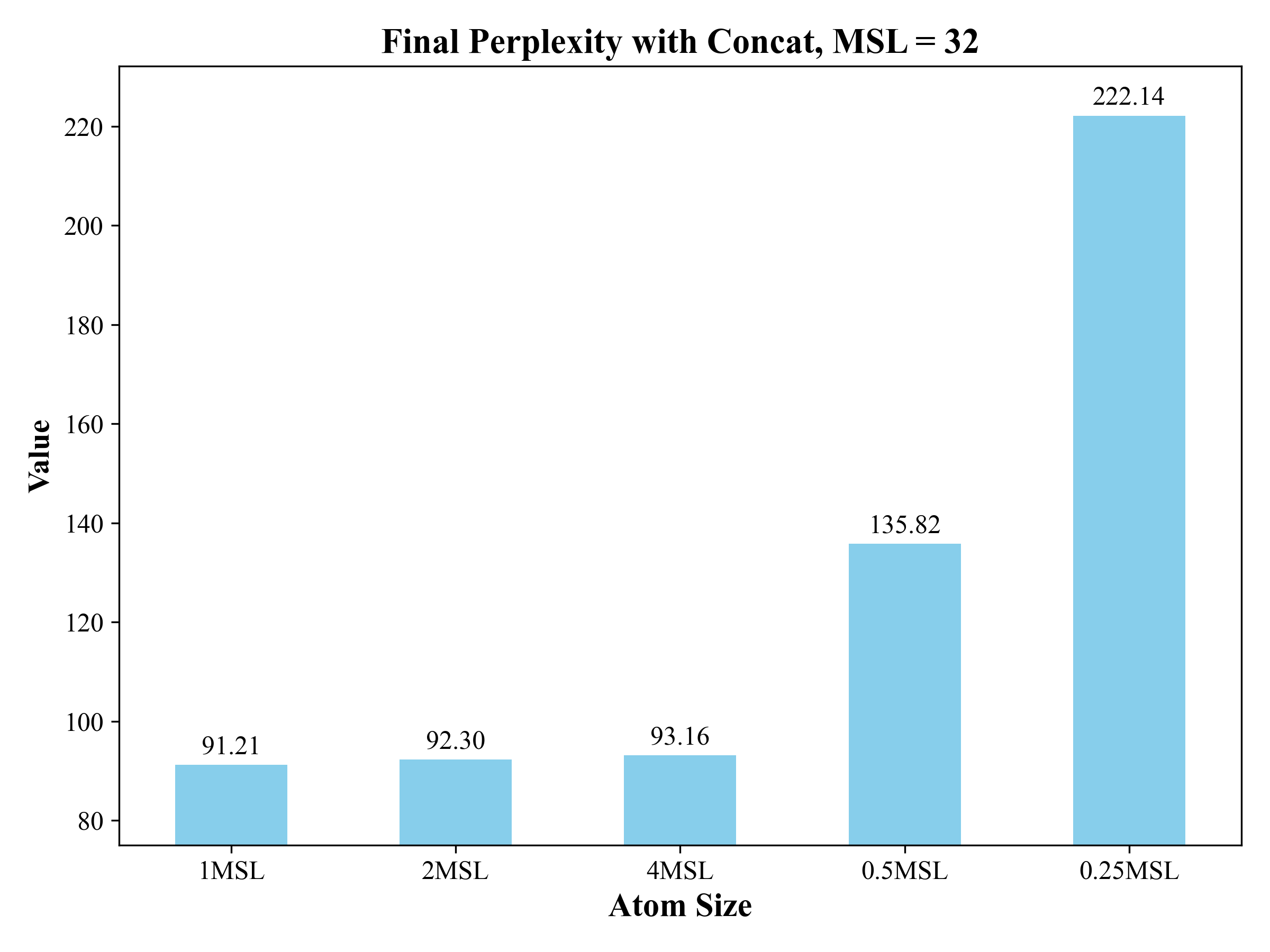}
        \caption{\textbf{Final Perplexity.} The model with atom size of 0.25MSL has the highest final perplexity value(222.14), while model with atom size of 1MSL has the lowest final perplexity value(91.21) for 2 epochs.}
    \end{subfigure}
    \hfill
    \begin{subfigure}[b]{0.45\textwidth}
        \includegraphics[width=\textwidth]{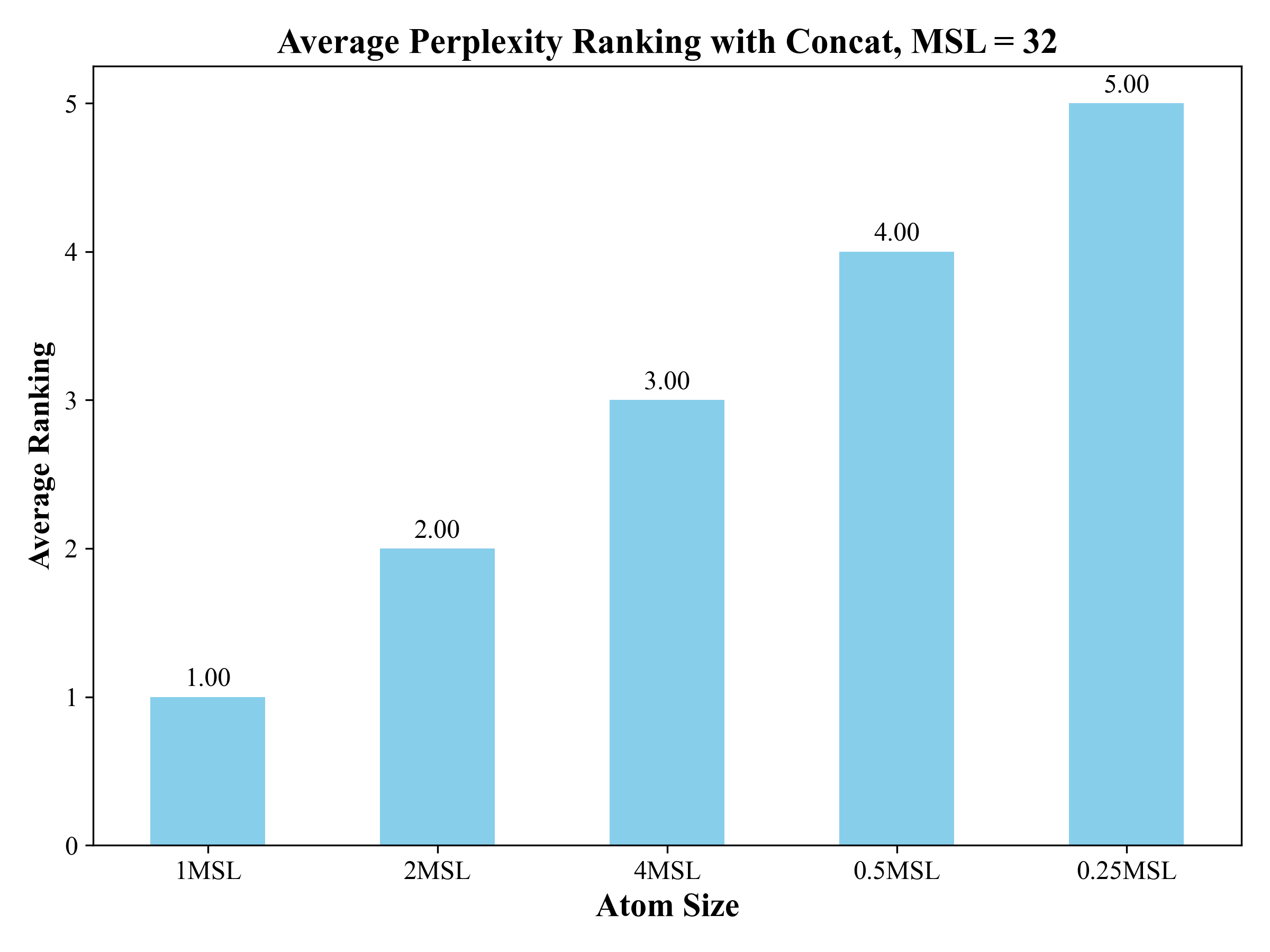}
        \caption{\textbf{Perplexity Ranking.}The model with atom size of 0.25MSL has the highest perplexity ranking(5), while model with atom size of 1MSL has the lowest perplexity ranking(1) for 2 epochs.}
    \end{subfigure}
    \caption{Comparisons across concat models with different atom sizes when MSL is 32. Smaller or larger atom sizes than 1MSL increase perplexity. The model with 1MSL as the atom size has the lowest final perplexity value and the smallest average perplexity ranking at the end of 2 epochs,  indicating better performance.}
    \label{fig:concat_32}
\end{figure*}

\begin{figure*}[t]
    \centering
    \begin{subfigure}[b]{0.45\textwidth}
        \includegraphics[width=\textwidth]{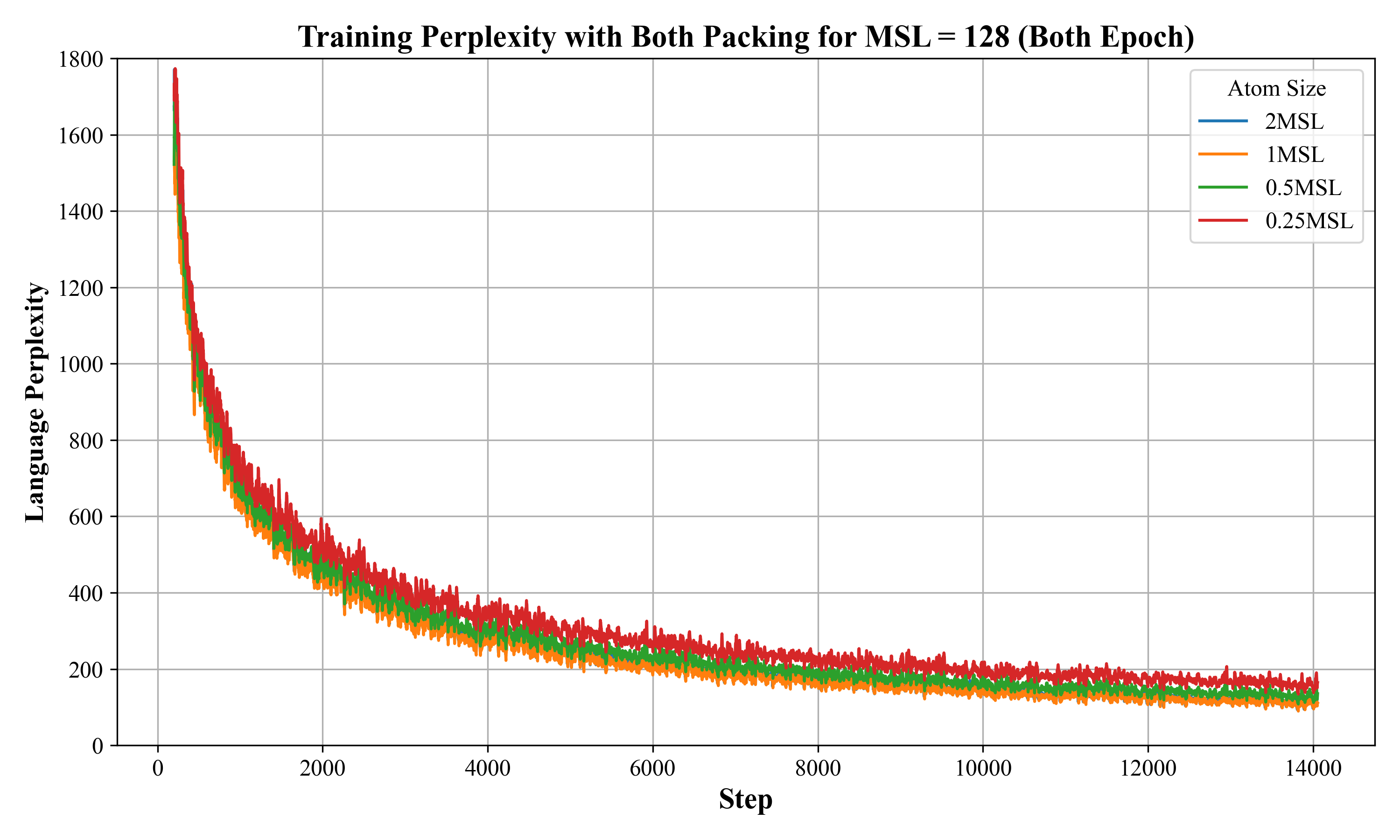}
        \caption{\textbf{Full Training Perplexity.} The models with atom sizes of 0.25MSL (red) and 0.5MSL (green) have higher perplexity than the others. 1MSL (orange) stabilizes at a low perplexity after an initial drop.}
    \end{subfigure}
    \hfill
    \begin{subfigure}[b]{0.45\textwidth}
        \includegraphics[width=\textwidth]{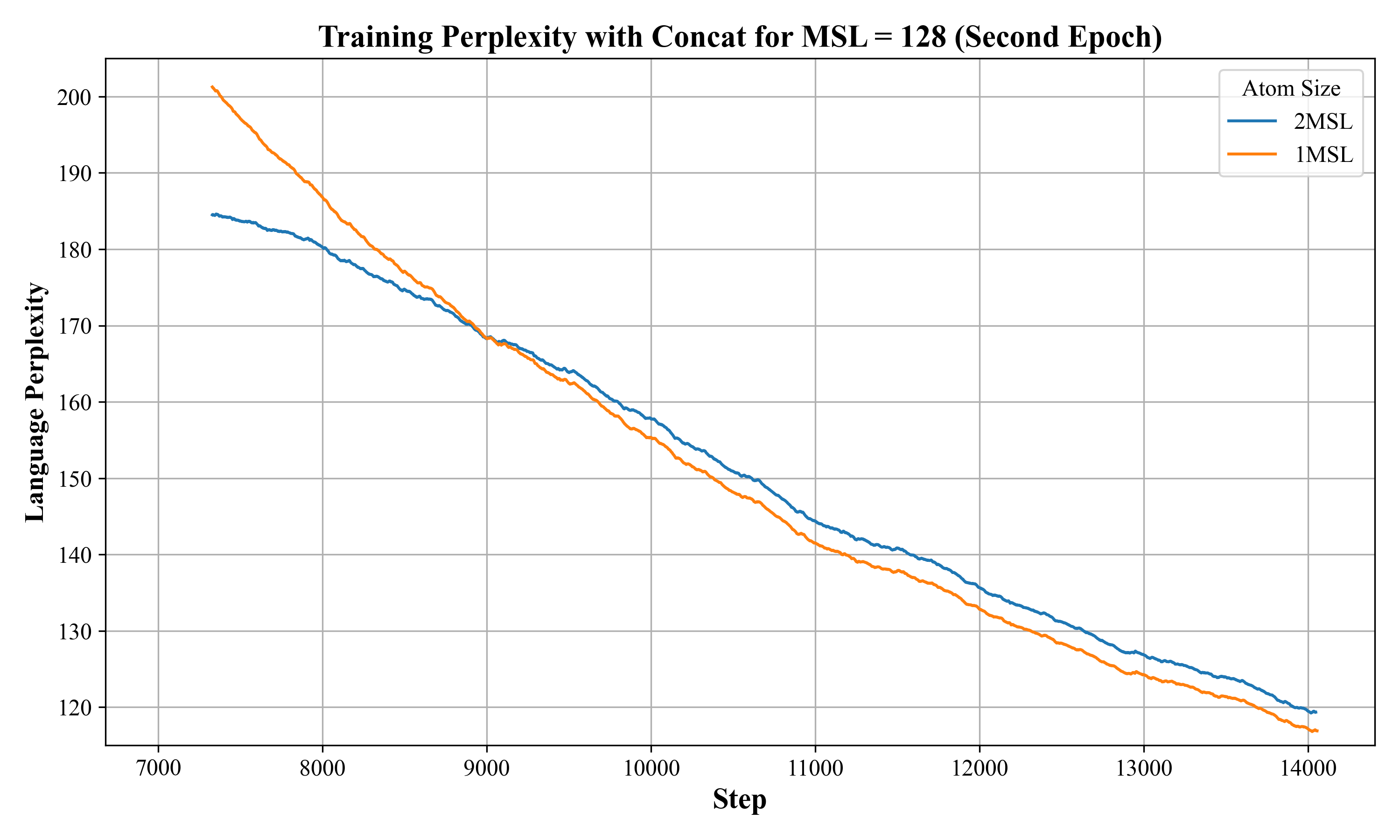}
        \caption{\textbf{Second Epoch Perplexity.} Initially, the model with atom size of 1MSL (orange) shows higher perplexity than 2MSL (blue). 1MSL continuously decreases and achieves the lowest perplexity by the end of the second epoch.}
    \end{subfigure}
    \vskip\baselineskip
    \begin{subfigure}[b]{0.45\textwidth}
        \includegraphics[width=\textwidth]{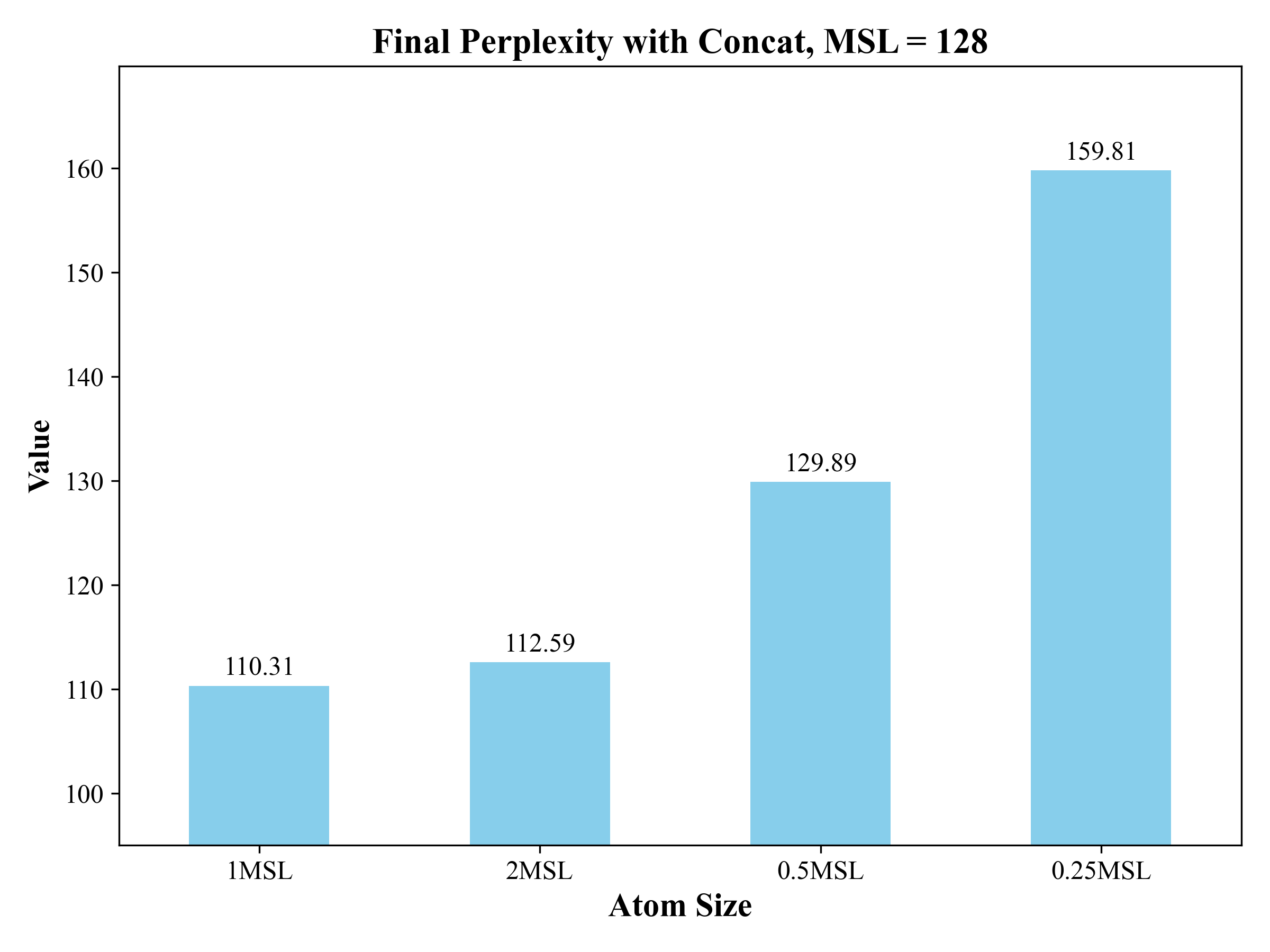}
        \caption{\textbf{Final Perplexity.} The model with atom size of 0.25MSL has the highest final perplexity value(159.81), while model with atom size of 1MSL has the lowest final perplexity value(110.31) for 2 epochs.}
    \end{subfigure}
    \hfill
    \begin{subfigure}[b]{0.45\textwidth}
        \includegraphics[width=\textwidth]{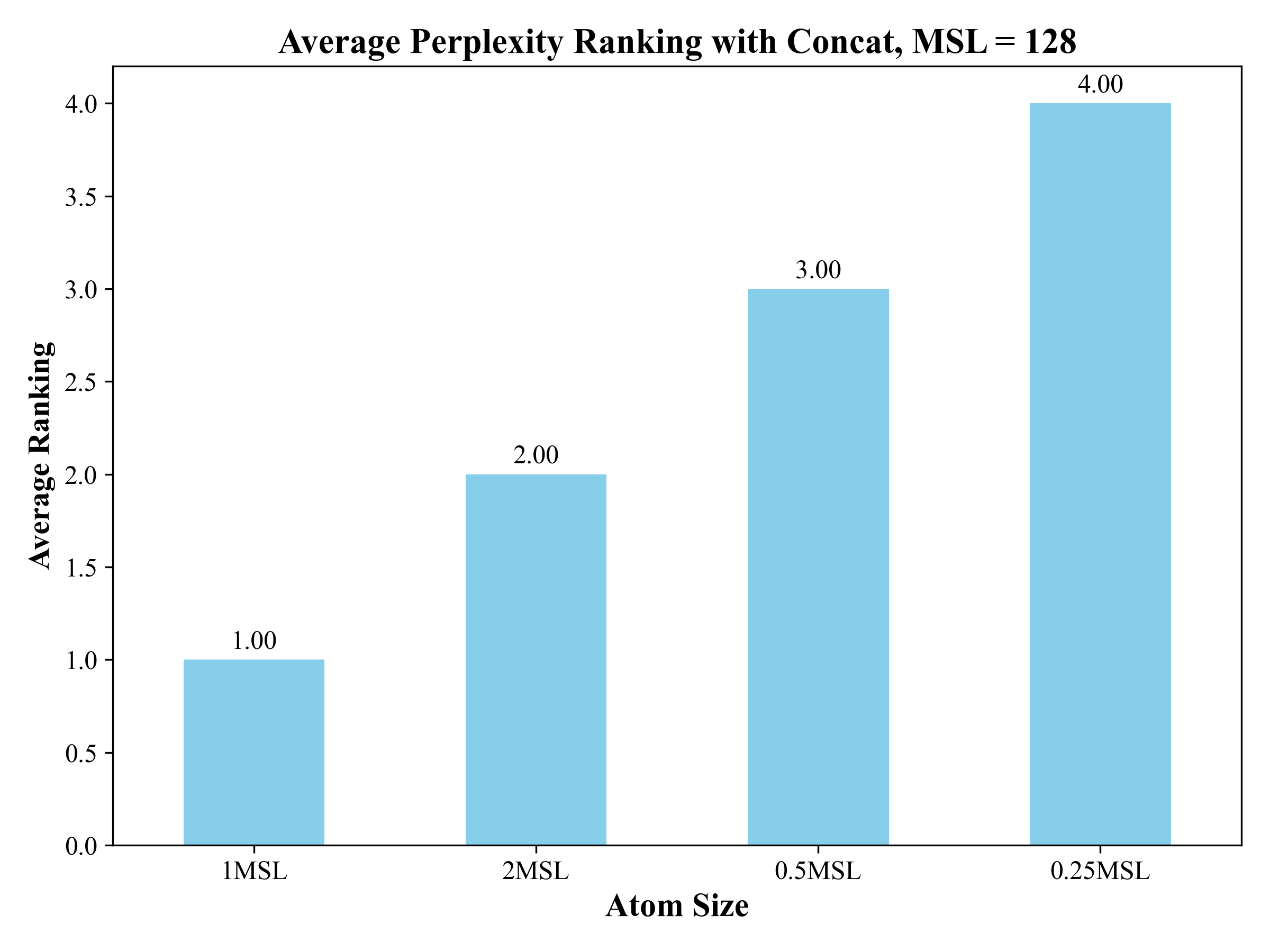}
        \caption{\textbf{Perplexity Ranking.}The model with atom size of 0.25MSL has the highest perplexity ranking (4), while model with atom size of 1MSL has the lowest perplexity ranking (1) for 2 epochs.}
    \end{subfigure}
    \caption{Comparisons across concat models with different atom sizes when MSL is 128. Smaller or larger atom sizes than 1MSL increase perplexity. The model with 1MSL as the atom size has the lowest final perplexity value and the smallest average perplexity ranking at the end of 2 epochs, indicating better performance.}
    \label{fig:concat_128}
\end{figure*}

\paragraph{\textbf{Concat.}}
See Figure~\ref{fig:concat_32}, Figure~\ref{fig:concat_128} for detail.

\begin{figure*}[t]
    \centering
    \begin{subfigure}[b]{0.45\textwidth}
        \includegraphics[width=\textwidth]{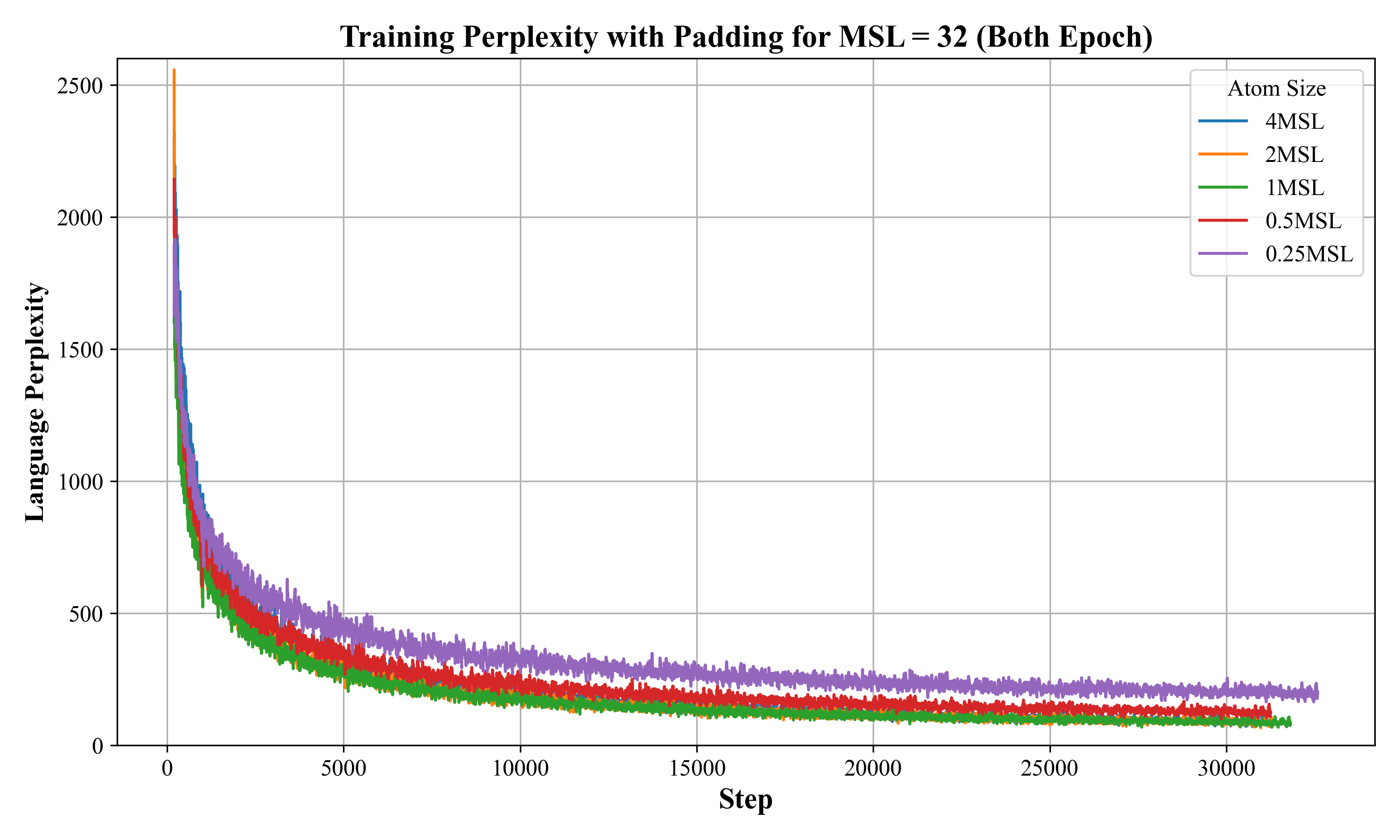}
        \caption{\textbf{Full Training Perplexity.} The models with atom sizes of 2MSL (orange) and 0.5MSL(red) have higher perplexity than the others. 1MSL (green) stabilizes at a low perplexity after an initial drop.}
    \end{subfigure}
    \hfill
    \begin{subfigure}[b]{0.45\textwidth}
        \includegraphics[width=\textwidth]{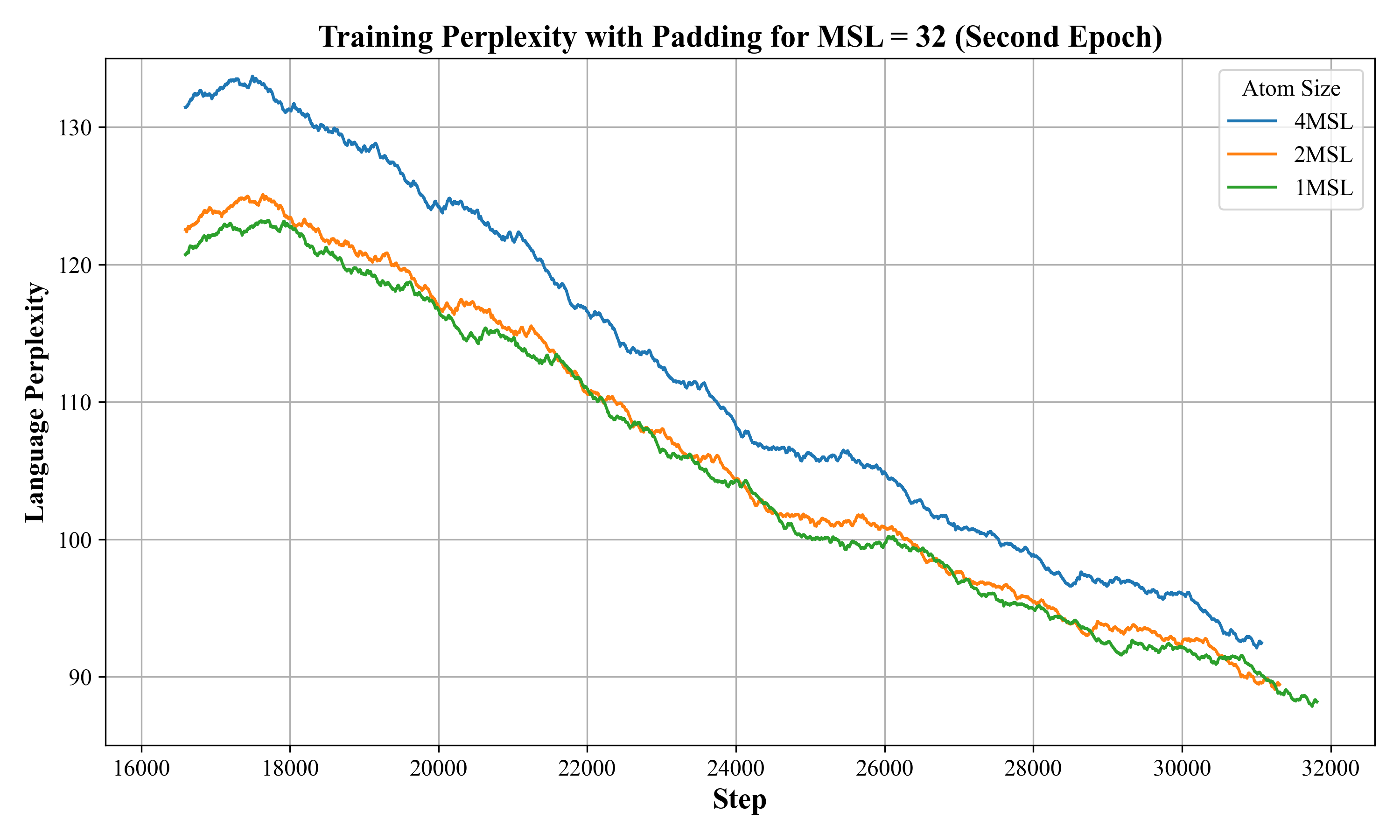}
        \caption{\textbf{Second Epoch Perplexity.} The models with atom size of 4MSL (blue) has higher perplexity than the other two. 1MSL (green) has the lowest perplexity at the end of second epoch.}
    \end{subfigure}
    \vskip\baselineskip
    \begin{subfigure}[b]{0.45\textwidth}
        \includegraphics[width=\textwidth]{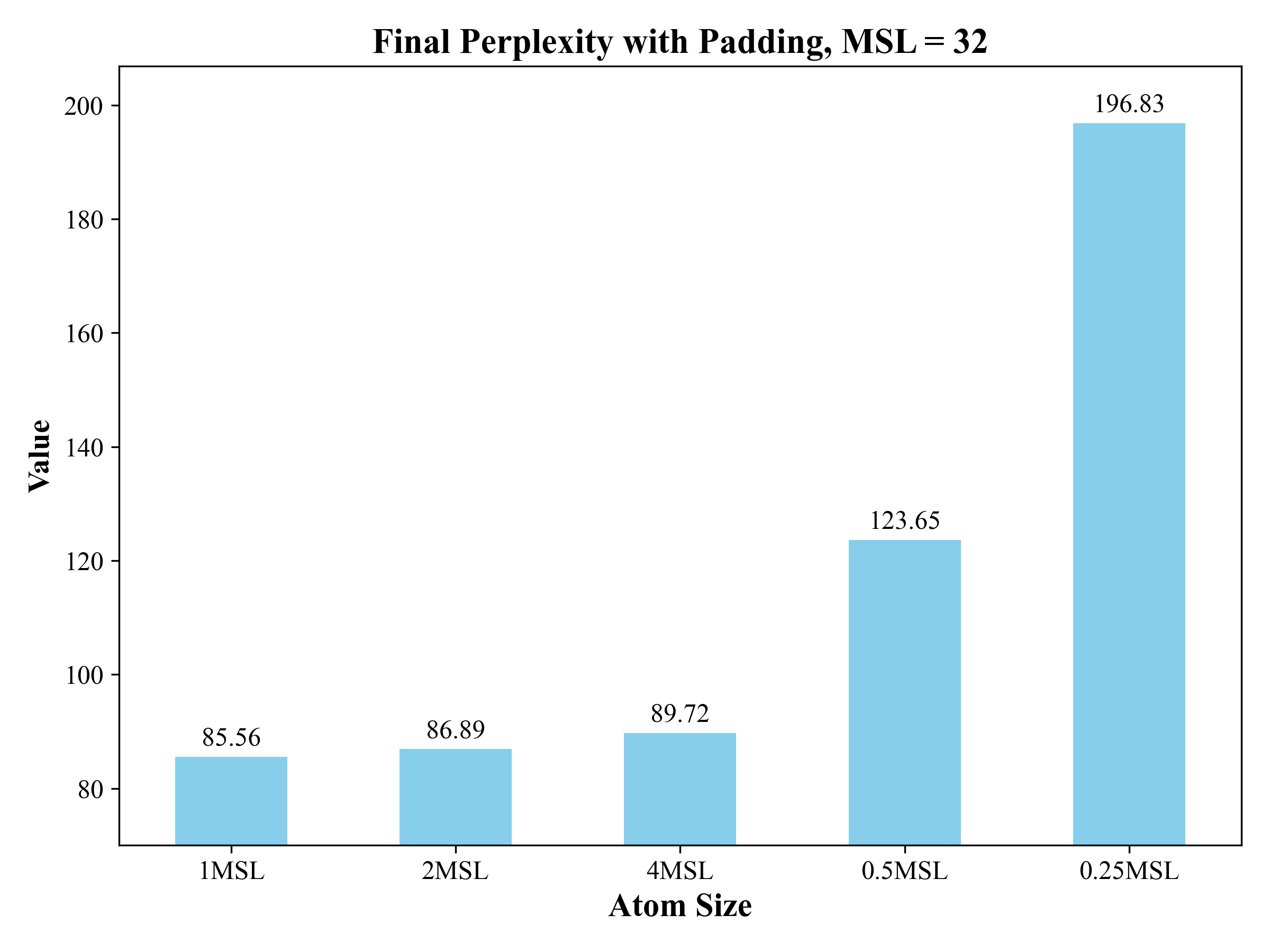}
        \caption{\textbf{Final Perplexity.} The model with atom size of 0.25MSL has the highest final perplexity value (196.83), while model with atom size of 1MSL has the lowest final perplexity value (85.56) for 2 epochs.}
    \end{subfigure}
    \hfill
    \begin{subfigure}[b]{0.45\textwidth}
        \includegraphics[width=\textwidth]{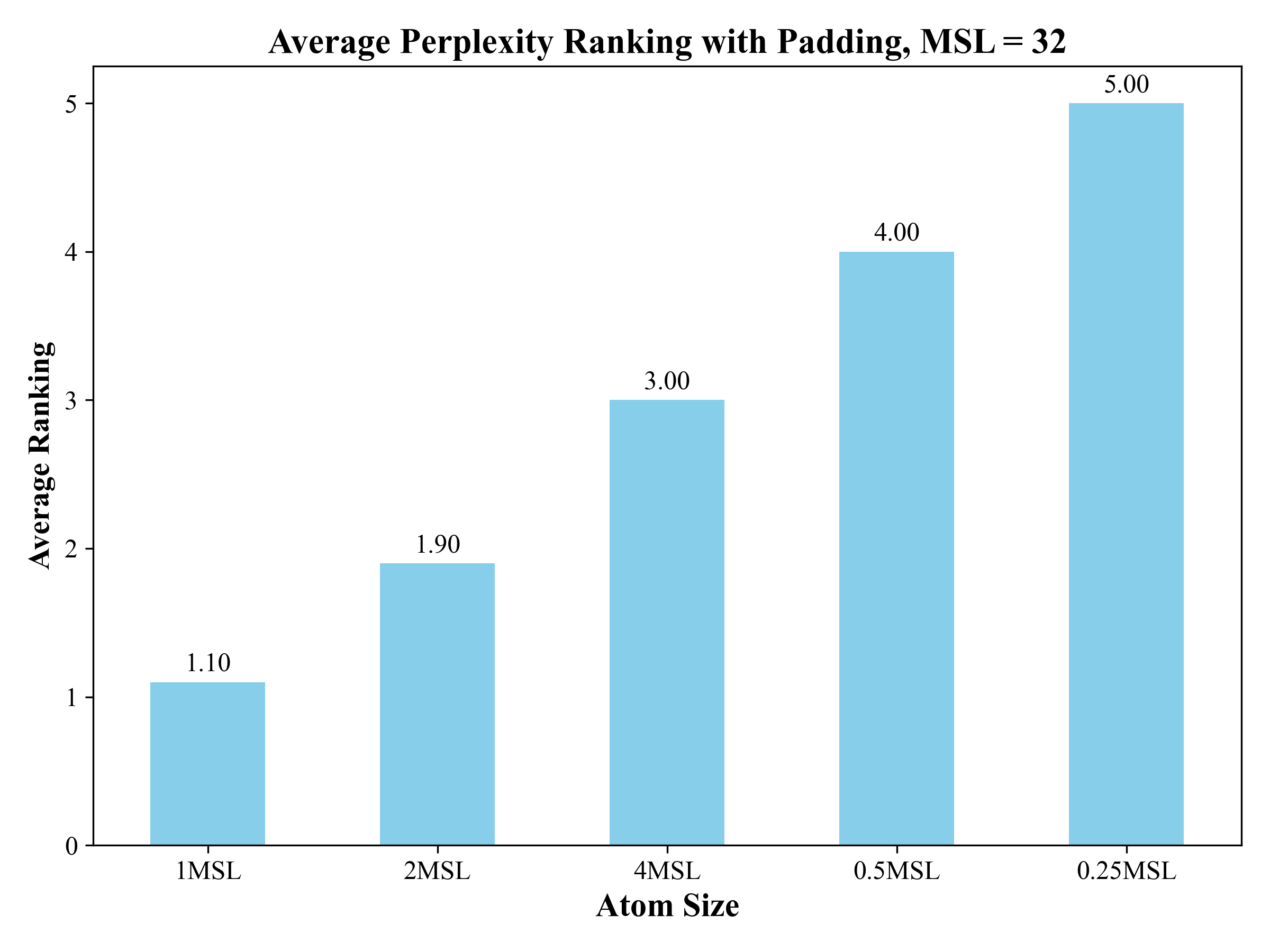}
        \caption{\textbf{Perplexity Ranking.} The model with atom size of 0.25MSL has the highest perplexity ranking (5), while model with atom size of 1MSL has the lowest perplexity ranking (1.1) for 2 epochs.}
    \end{subfigure}
    \caption{Comparisons across padding models with different atom sizes when MSL is 32.Smaller or larger atom sizes than 1MSL increase perplexity. The model with 1MSL as the atom size has the lowest final perplexity value and the smallest average perplexity ranking at the end of 2 epochs,  indicating better performance.}
    \label{fig:padding_32}
\end{figure*}

\begin{figure*}[t]
    \centering
    \begin{subfigure}[b]{0.45\textwidth}
        \includegraphics[width=\textwidth]{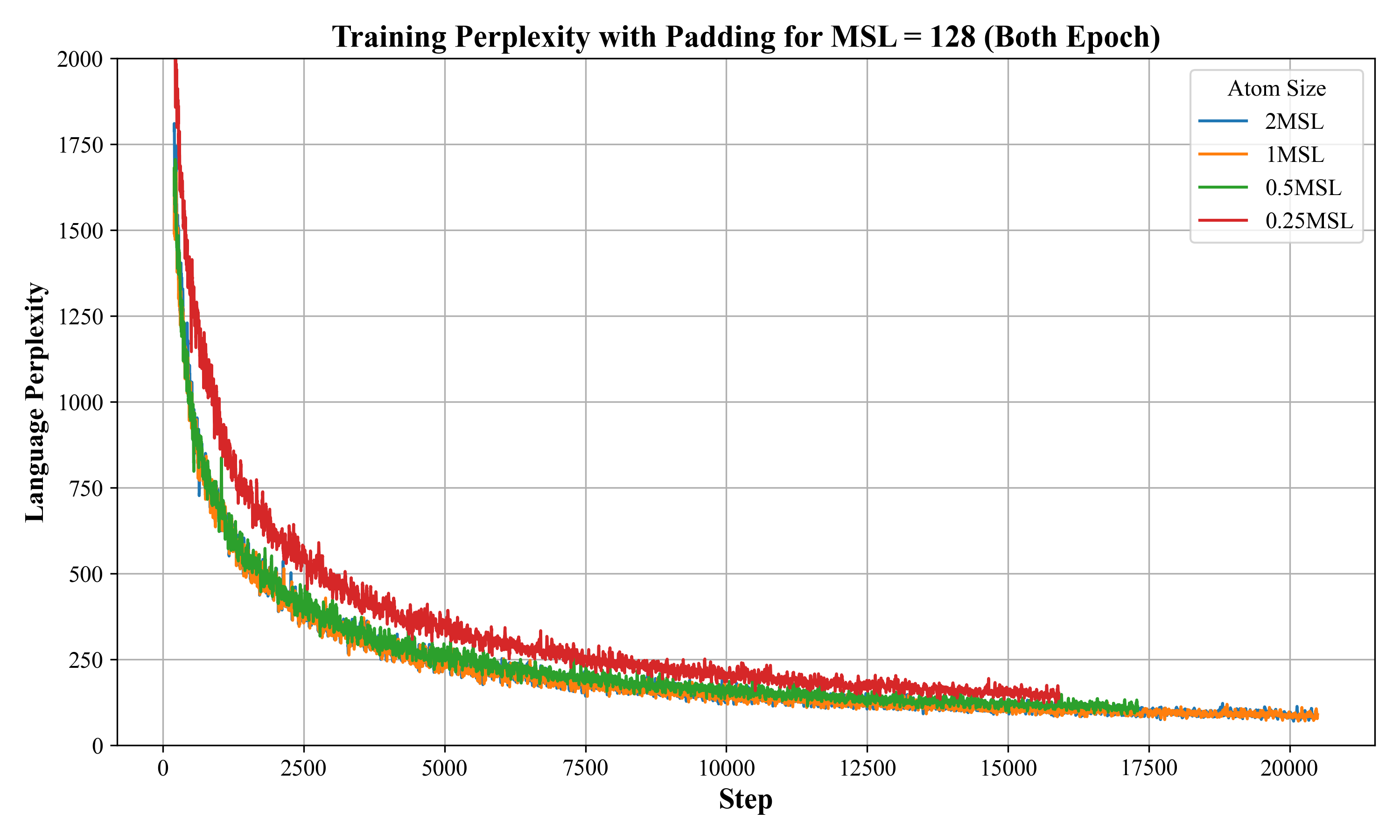}
        \caption{\textbf{Full Training Perplexity.} The model with atom sizes of 0.25MSL (red) has higher perplexity than the others. 1MSL (orange) stabilizes at a low perplexity after an initial drop.}
    \end{subfigure}
    \hfill
    \begin{subfigure}[b]{0.45\textwidth}
        \includegraphics[width=\textwidth]{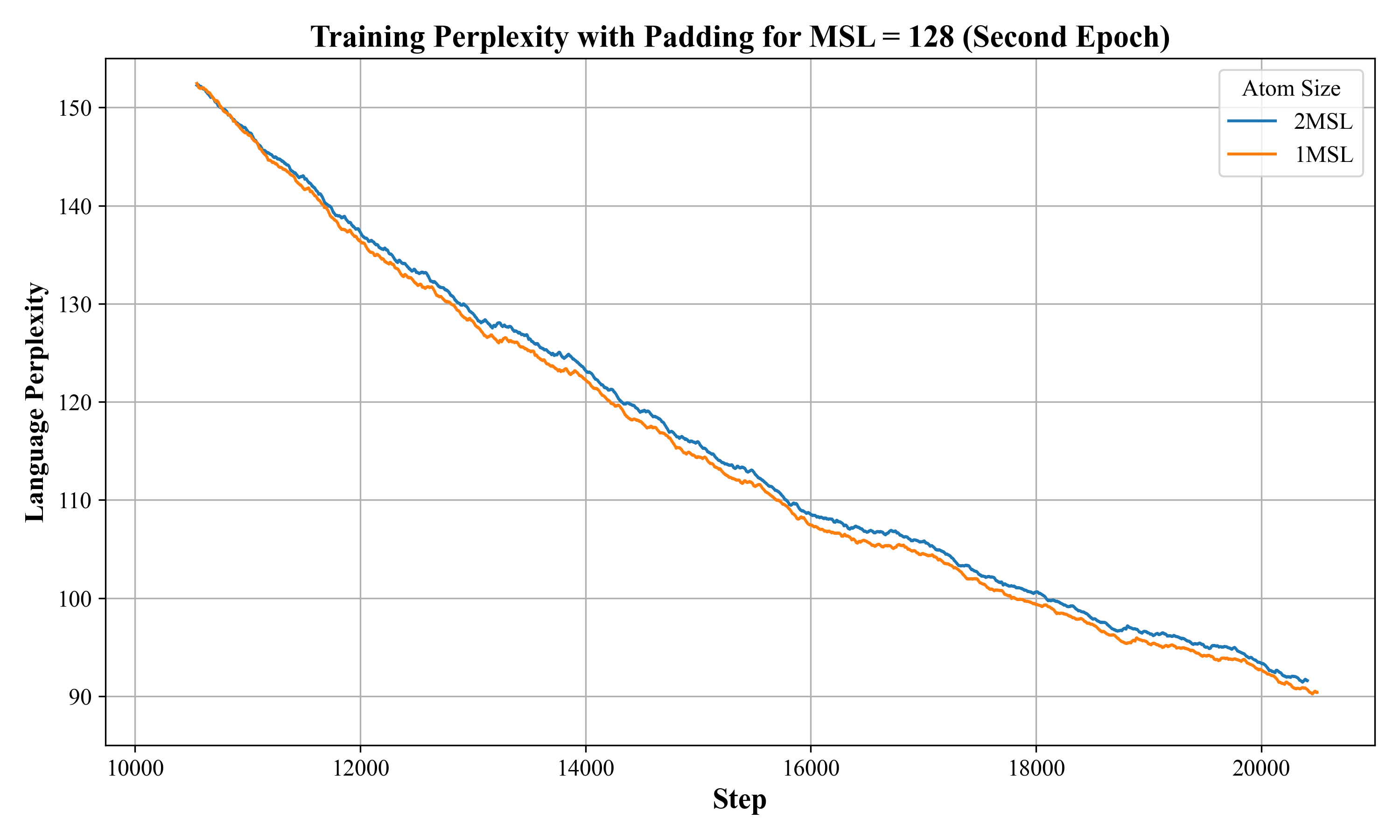}
        \caption{\textbf{Second Epoch Perplexity.} Initially, the model with atom size of 2MSL (blue) and 1MSL(orange) have similar perplexity. 1MSL (orange) has the lowest perplexity at the end of second epoch.}
    \end{subfigure}
    \vskip\baselineskip
    \begin{subfigure}[b]{0.45\textwidth}
        \includegraphics[width=\textwidth]{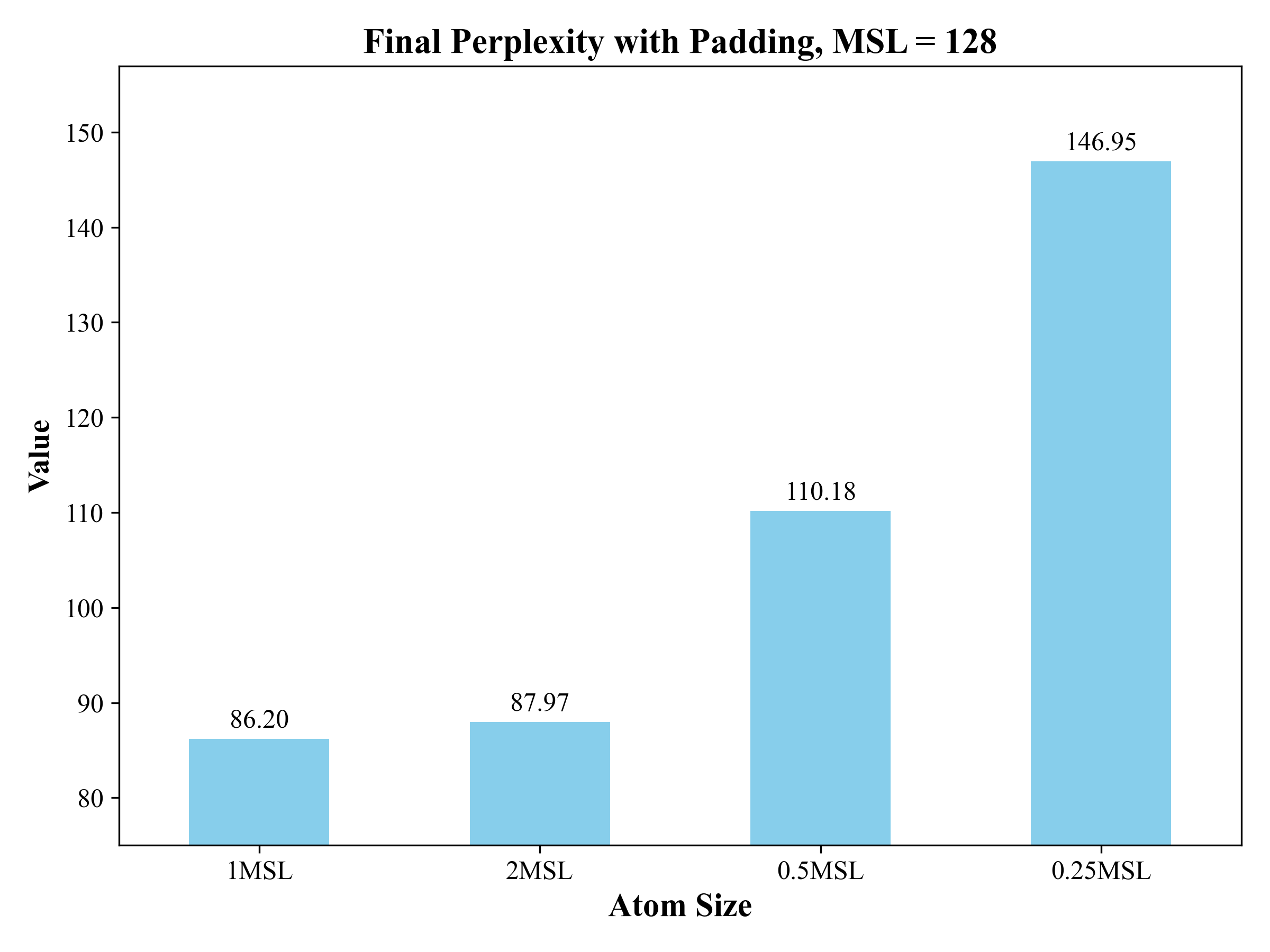}
        \caption{\textbf{Final Perplexity.} The model with atom size of 0.25MSL has the highest final perplexity value (146.95), while model with atom size of 1MSL has the lowest final perplexity value (86.20) for 2 epochs.}
    \end{subfigure}
    \hfill
    \begin{subfigure}[b]{0.45\textwidth}
        \includegraphics[width=\textwidth]{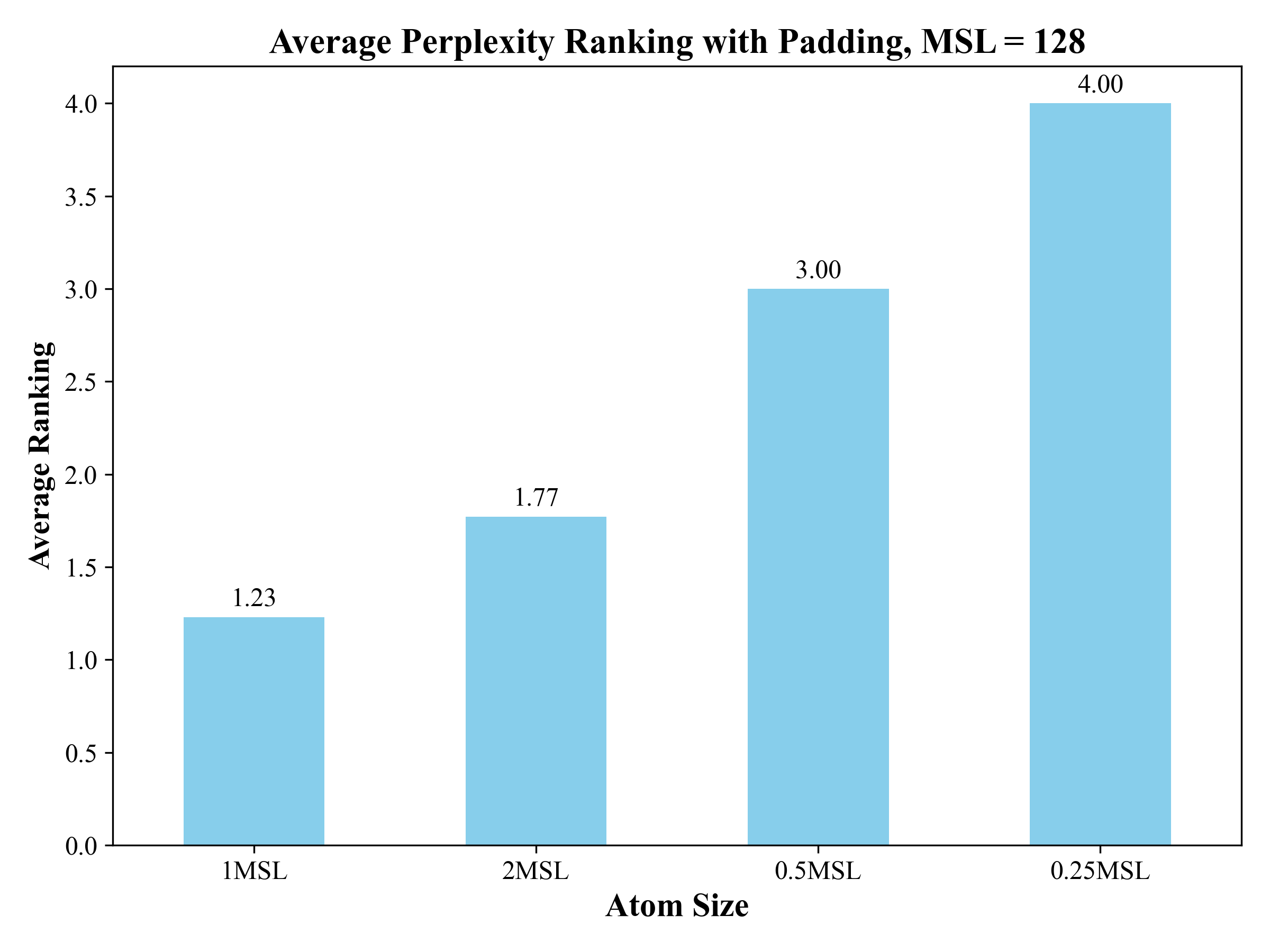}
        \caption{\textbf{Perplexity Ranking.}The model with atom size of 0.25MSL has the highest perplexity ranking (4), while model with atom size of 1MSL has the lowest perplexity ranking (1.23) for 2 epochs.}
    \end{subfigure}
    \caption{Comparisons across padding models with different atom sizes when MSL is 128. Smaller or larger atom sizes than 1MSL increase perplexity. The model with 1MSL as the atom size has the lowest final perplexity value and the smallest average perplexity ranking at the end of 2 epochs,  indicating better performance.}
    \label{fig:padding_128}
\end{figure*}

\paragraph{\textbf{Padding.}}
See Figure~\ref{fig:padding_32}, Figure~\ref{fig:padding_128} for detail.

\subsection{Step Size Differences in Padding }\label{sec:step_size_diff_in_padding}
As mentioned in Section~\ref{sec:padding_experiment}, models with different atom sizes have different training steps due to the differing amounts of padding tokens in the training sequences. In specific, we added padding tokens to the dataset in two ways.

\textbf{(a) By the End of Each Subsequence.} As mentioned in Appendix~\ref{sec:concat_and_padding}, documents are split to subsequences with an <EOS> token added to their ends. In our case, the <EOS> is the same as the padding token.

\textbf{(b) By the End of Tails.} During the splitting process, we produce some end tails which do not completely fill an atom size as shown in figure~\ref{fig:padding}. Padding tokens are added to those tails to make sure that they have lengths equal to the atom size.

Smaller atom sizes will have more padding tokens from source (a) than larger atom sizes because they have larger numbers of subsequences. Larger atom sizes will have more padding tokens from source (b) because we need more padding tokens for the end tails to fulfill the length requirement.

\subsection{License for GPT-2} 
GPT-2 used a Modified MIT License, which can be seen on this: \url{https://github.com/openai/gpt-2/blob/master/LICENSE}. We only use GPT-2 for research, which is consistent to its intended use.

\end{document}